\documentclass[journal,twoside,web]{ieeecolor}
\usepackage{jsen}
\usepackage{cite}
\usepackage{textcomp}
\usepackage{wrapfig}
\usepackage{amssymb}
\usepackage{amsmath}
\usepackage{booktabs}
\usepackage{multirow}
\usepackage{longtable}
\usepackage{url}
\usepackage{threeparttable}
\usepackage{subfigure}
\usepackage{algorithm}  
\usepackage{adjustbox}
\usepackage{makecell}
\usepackage{algpseudocode}  
\usepackage{float}
\usepackage{amssymb}
\usepackage{graphicx}
\usepackage{amsmath}
\usepackage{booktabs}
\usepackage{multirow}
\usepackage{longtable}
\usepackage{url}
\usepackage{threeparttable}
\usepackage{subfigure}
\usepackage{algorithm}  
\usepackage{adjustbox}
\usepackage{makecell}
\usepackage{algpseudocode}  
\usepackage{orcidlink}
\usepackage{float}

\def\BibTeX{{\rm B\kern-.05em{\sc i\kern-.025em b}\kern-.08em
    T\kern-.1667em\lower.7ex\hbox{E}\kern-.125emX}}
\definecolor{abstractbg}{rgb}{0.89804,0.94510,0.83137}
\begin{document}
\title{A Dual Attentive Generative Adversarial Network for Remote Sensing Image Change Detection}
\author{Luyi~Qiu \orcidlink{0000-0002-8158-4604},
        Xiaofeng Zhang \orcidlink{0000-0002-7185-4682}
        ChaoChen Gu*, ~\IEEEmembership{Member,~IEEE, }
        and ShanYing Zhu, ~\IEEEmembership{Member,~IEEE }
\thanks{Manuscript received July 20, 2023. This work was supported by Joint
Fund of the Ministry of Education No.8091B022101}
\thanks{LuYi Qiu, XiaoFeng Zhang, ChaoChen Gu, ShanYing Zhu are now with School of Electronic Information and Electrical Engineering, Shanghai Jiao Tong University ((Corresponding author: Chaochen Gu, his email: jacygu@sjtu.edu.cn)}
}

\IEEEtitleabstractindextext{%
\fcolorbox{abstractbg}{abstractbg}{%
\begin{minipage}{\textwidth}%
\begin{wrapfigure}[13]{r}{3in}%
\includegraphics[width=2.8in]{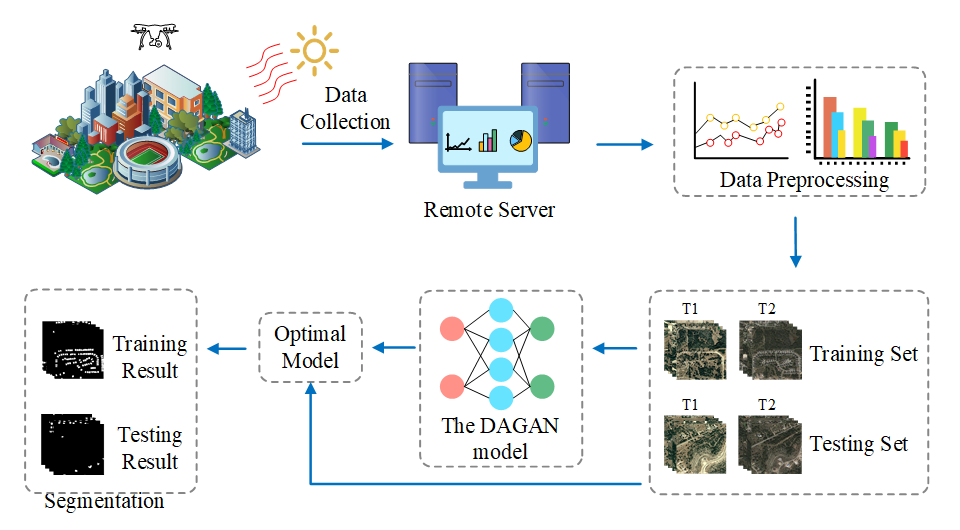}%
\end{wrapfigure}%
\begin{abstract}

Remote sensing change detection between bi-temporal images receives growing concentration from researchers. However, comparing two bi-temporal images for detecting changes is challenging, as they demonstrate different appearances. In this paper, we propose a dual attentive generative adversarial network for achieving very high-resolution remote sensing image change detection tasks, which regards the detection model as a generator and attains the optimal weights of the detection model without increasing the parameters of the detection model through generative-adversarial strategy, boosting the spatial contiguity of predictions. Moreover, We design a multi-level feature extractor for effectively fusing multi-level features, which adopts the pre-trained model to extract multi-level features from bi-temporal images and introduces aggregate connections to fuse them. To strengthen the identification of multi-scale objects, we propose a multi-scale adaptive fusion module to adaptively fuse multi-scale features through various receptive fields and design a context refinement module to explore contextual dependencies. Moreover, the DAGAN framework utilizes the 4-layer convolution network as a discriminator to identify whether the synthetic image is fake or real. Extensive experiments represent that the DAGAN framework has better performance with 85.01\% mean IoU and 91.48\% mean F1 score than advanced methods on the LEVIR dataset.
\end{abstract}

\end{minipage}}}

\maketitle

\section{Introduction}
As a significant direction of image processing, change detection for very high-resolution remote sensing (VHR) images are applied in automatic driving, precision agriculture, and damage assessment\cite{79}, which aims for locating and segmenting the surface changes from the co-registered pair of images captured at different moments. 

Early traditional change detection methods, such as fuzzy c-mean \cite{69}, support vector machines \cite{70}, and random forest \cite{71}, mainly rely on hand-crafted features designed by experienced experts in the remote sensing field to obtain change maps. However, these hand-crafted features with personal preferences usually lead to the accuracy limitation of change detection. Compared with traditional methods, convolutional neural networks (CNNs) with excellent representation capability could automatically learn significant features from massive images through the training period and have been applied in many image processing tasks, like semantic segmentation \cite{79,80}, object detection \cite{67}, image reconstruction \cite{68}, etc.

Some scholars regard the change detection task as a segmentation task, which segments the changes by classifying the bi-temporal features extracted by convolutional neural networks into different categories \cite{51,52,53,81,82,83}. For instance, Chen et al. \cite{51} converted bi-temporal images into tokens and adopted a novel transformer to model the spatial-temporal domain contexts. Shi et al. \cite{52} designed a metric module to evaluate the importance of features, utilized attention modules to obtain more valuable features, and adopted a supervised module as an auxiliary to increase the recognization ability. Yang et al. \cite{53} located change regions through feature pairs captured from different modules and applied other parameters to explore the discrepancy across land-cover distributions.

With the sensors' development, the RS images' resolution has increased, which could contain multi-scale spatial details of objects, and object categories have an unbalanced distribution. However, simple change detection methods cannot be accomplished by comparing two bi-temporal images since they demonstrate different appearances. Based on the above observation, we propose a novel DAGAN framework for change detection tasks, which relieves problems through the following three aspects. \textbf{1) Multi-level feature extractor.} The DAGAN framework adopts the pre-trained model to extract multi-level features from bi-temporal images in a coarse-to-fine strategy and introduces aggregate connections to fuse multi-level features progressively.  \textbf{2) Multi-scale adaptive fusion module.} The DAGAN framework uses a multi-scale adaptive fusion module to adaptively fuse multi-scale features through various receptive fields and make the framework focus on semantic features. \textbf{3) Context refinement module.} The DAGAN framework explores contextual dependencies on the global region through context refinement modules.

Compared with the forward propagation of CNNs, GAN reason and predict change regions through generative adversarial learning \cite{7}, which could assist predictions have more spatial contiguity \cite{22}. Some studies have explored applications in image processing fields through generative adversarial networks \cite{19,20}. The DAGAN framework optimizes the model's weights through a generative-adversarial strategy, strengthening the spatial contiguity of predictions without increasing the number of parameters.

Above all, the contributions are as follows:

\begin{enumerate}
\item We propose a novel dual attentive generative-adversarial framework (DAGAN) for achieving accurate change detection, which regards a dual attentive network (DANet) as a generator and attains the optimal weights of the detection model without increasing the parameters of the detection model through generative-adversarial strategy, boosting the spatial contiguity of predictions.
\item We propose a multi-scale adaptive fusion module to adaptively fuse multi-scale features through various receptive fields and design a context refinement module to explore contextual dependencies, which could assist the framework in enhancing the recognition ability of multi-scale objects.
\item Extensive experiments prove that the DAGAN framework performs better than advanced methods on the LEVIR datasets. 
\end{enumerate}

\section{Related Work}
\subsection{Remote Sensing Change Detection}

Recently, numerous efforts have been devoted to improving advanced architectures to achieve multi-level feature fusion for gaining promising change detection results \cite{54,56,72,73,74,75,76}. For instance, Daudt et al. \cite{72} attempted to use the adding operation, a concatenation operation, or a concatenation operation based absolute value of features to fuse multi-level features from bi-temporal, respectively. Chen et al. \cite{73} calculated the attention weights between two pixels from paired images through the self-attention mechanism, which assists the framework in improving the recognition ability of differences. Papadomanolaki et al. \cite{74} used LSTM blocks to model the spatial relationship of multi-level features extracted from paired images. Fang et al. \cite{75} refined difference features extracted from multi-level layers through ensemble channel attention modules. Zhang et al. \cite{76} enhanced boundary completeness of multi-level features by adopting attention modules to fuse differences. Li et al. \cite{56} constructed a neighbor aggregation module to combine multi-level features, captured temporal differences from paired images through a progressive change identifying module, and reweighted features from high levels to low levels through a supervised attention module.

Meanwhile, VHR images contain multi-scale details of objects, and objects have class-imbalance problems \cite{54,56,72,73,74,75,76}. Some scholars tried to improve multi-scale fusion modules \cite{90,91,92}, and architectures \cite{86, 87, 80} attempted to improve the recognition ability. For instance, Yang et al. \cite{90} aggregated multi-scale difference features extracted from images at T1 and T2 moment through fusion blocks, which could generate a framework with strong separability.  Lv et al. \cite{92} embedded multiscale convolution modules into a UNet, which could assist the framework to capture various sizes and shapes of objects. However, these methods only implicitly capture global relationships through multiple convolution operations and lack a global dependencies of images, which cannot build dependencies between features and fails to use global contextual information fully. Wang et al. \cite{86} proposed an improved self-attention module in which the effectiveness of non-local operations on spatial-temporal dimensions of videos and images was mainly explored. 

Motivated by these, the DAGAN designs multi-scale adaptive fusion modules for progressively aggregating multi-scale feautres through various receptive fields from multi-level features extracted by pre-trained models in a coarse-to-fine strategy and uses context refinement module for exploring contextual relationship and capturing long dependencies on the global region. 

\subsection{Generative Adversarial Network}

The GAN comprises a generator and a discriminator, where the generator collects the distribution of the dataset, and the discriminator assesses the probability that data is real/fake. Compared with the forward propagation of CNNs, generative adversarial networks could assist predictions of the framework with more spatial contiguity. Generative adversarial networks estimate the optimal weights of models through generative adversarial learning, which makes the generator synthesize data to confuse the discriminator and causes it to be unable to distinguish whether it is fake or real, which could achieve dynamic balance and generate various synthetic images through continuous learning and confrontation of the generator and discriminator \cite{7}.

Some scholars have successfully applied GANs to the remote sensing images of dense prediction tasks, such as segmentation \cite{19,20}, and reconstruction \cite{65}, which prediction models as generators and optimize weights of models through generative-adversarial strategy without increasing the parameters of prediction models, making prediction have more spatial contiguity. For instance, Hamghalam et al. \cite{19} adopted the Cycle-GAN framework with attention mechanisms to improve the contrast within underlying structures. Zhang et al. \cite{20} designed a novel GAN with dual Markovian discriminators, which could fully preserve image features. Niu et al. \cite{65} synthesized HDR from multi-exposed LDR images through a generative adversarial network, which generated reliable information in the missing region without introducing artifacts, utilized a reference-based block to align features, and adopted a deep supervision strategy for eliminating reconstruction artifacts. And it could be found that the generative adversarial strategy could effectively improve the optimism processing to improve the performance.

Inspired by literature, the DAGAN framework optimizes the model's weights through a generative-adversarial strategy, which aims to assist predictions of DANet to get more spatial contiguity.

\section{Method}
\subsection{Overall Architecture}

\begin{figure}[htbp]
    \centering \includegraphics[width=\linewidth]{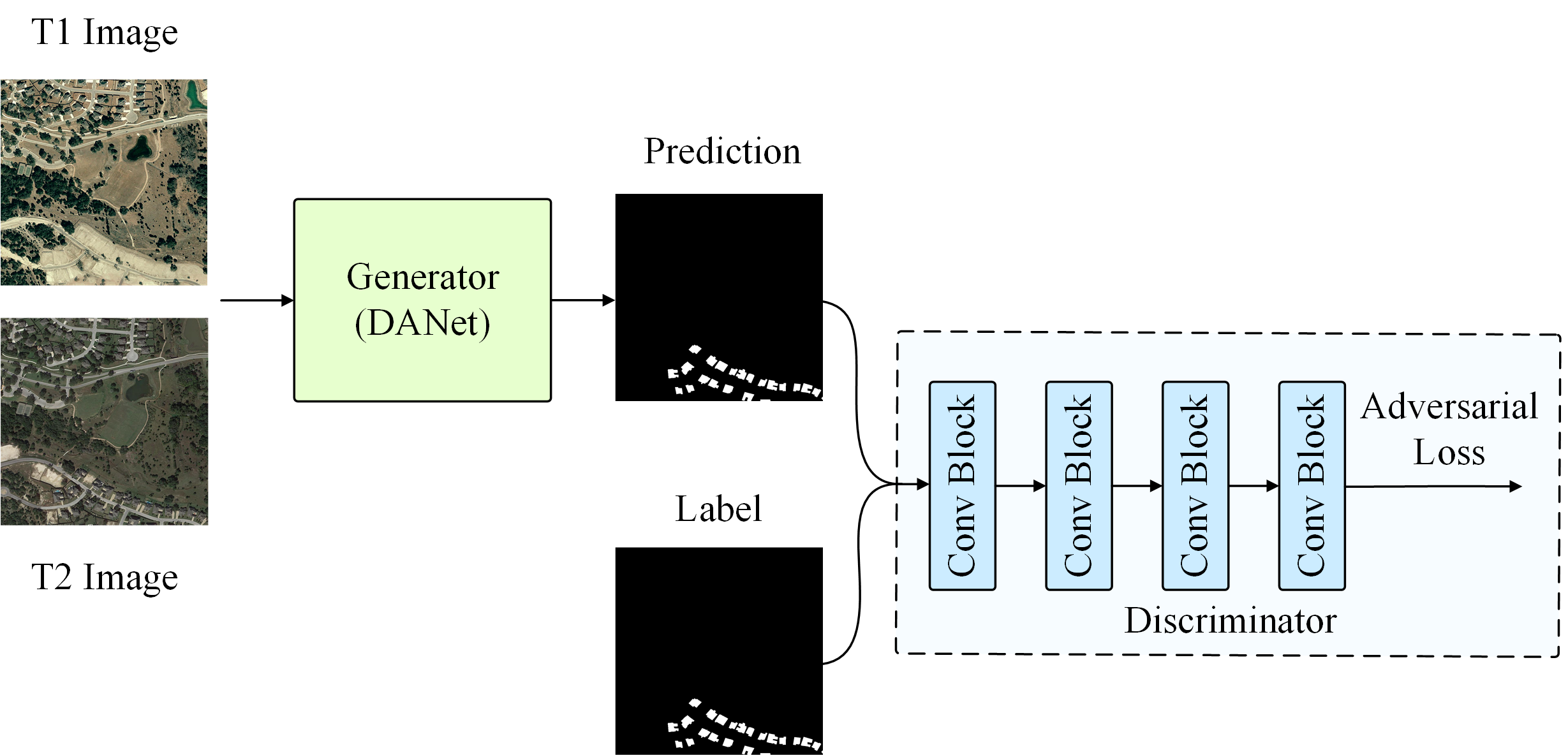}
	\caption{The overall architecture of the DAGAN framework.\label{fig0}}
\end{figure}

The DAGAN framework is shown in Figure \ref{fig0}, which mainly consists of a generator $G$ and a discriminator $D$. 

The DAGAN framework designs a dual attentive network (DANet) as a generator to capture multi-level features of paired images from different moments and generate predictions of change detection. After that, the DAGAN framework inputs the prediction into the 4-layer discriminator. The discriminator D evaluates whether the prediction is fake or real, representing differences between the prediction and the ground truth. And DAGAN framework utilizes a generative-adversarial strategy to optimize the weights of the framework, which mainly relies on the deep supervision of the generative loss and the adversarial loss. Furthermore, the optimism processing as follows in Algorithm \ref{alg1}.

 \begin{algorithm}[h]  
	\caption{The optimism processing via the DAGAN. \label{alg1}}  
	\begin{algorithmic}[1]  
		\Require Training set
		\Ensure Change detection result.
		\State Train the DAGAN framework with the LEVIR training set and predict the results of change detection tasks.
            \For{ each class: }
		\State Design class pairs via the confusion matrix.
            \For{$m$ steps do: }
                \State Update the generator via $L_G$ loss.
                \State Update the discriminator via $L_D$ loss.
		\EndFor
            \State Train the generator to generate the synthetic results.
            \State Add the synthetic results to the LEVIR training set.
		\EndFor 
        \State Train the DAGAN framework with the new dataset to predict the final results.
	\end{algorithmic}  
\end{algorithm}

\subsection{Dual Attentive Network}

\begin{figure*}[htbp]
    \centering
    \includegraphics[width=18cm]{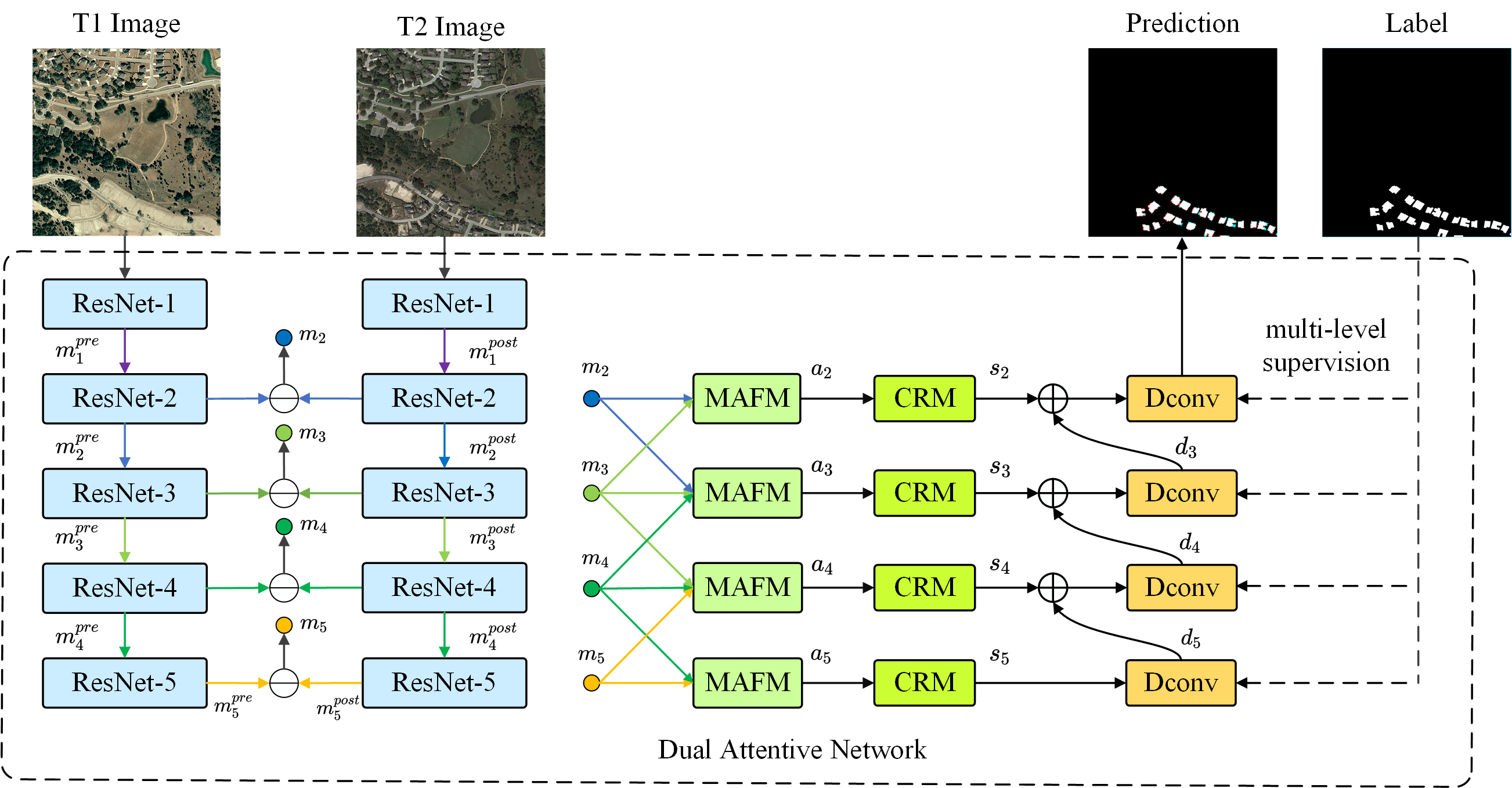}
    
	\caption{The overview of the DAGAN framework, where the MAFM module represents the multi-Scale adaptively fusion module, and the CRM module indicates the context refinement module.\label{fig2}}
\end{figure*}

The dual attentive network (DANet) is constructed upon an encoder-decoder architecture, as Figure \ref{fig2} shows. The DANet firstly extracts multi-level features from paired images at T1 and T2 moments through two pre-trained models. For better-adapting change detection tasks, the DANet removes the global average pooling and the fully connected layer of the pre-trained models and adopts the convolution with a stride size 2 in each stage of the pre-trained models to downsample bi-temporal image features, as Equation \ref{eq2} and \ref{eq3}.

\begin{equation}
    \label{eq2}
	m_{i}^{pre} = pre_{i}(x_{pre}), 
\end{equation}
where $x^{pre}$ denotes the image at the T1 moment, $x^{post}$ indicates the image at the T2 moment, $ pre(\cdot)$ indicates the pre-trained model and $i$ represents the stage of the pre-trained model.
\begin{equation}
 \label{eq3}
	m_{i}^{post} =pre_{i}(x_{post}), 
\end{equation}

Then, DANet integrates features captured from the nearby stages of the pre-trained model through add operations, as Equation \ref{eq4}, \ref{eq5}, and \ref{eq6}.

\begin{equation}
\label{eq4}
 m_{2} = pre_{2}(x)+pre_{3}(x), 
\end{equation}
where $m_2$ aggregates features extracted from the twice and third stages of the pre-trained model.
\begin{equation}
\label{eq5}
m_{i} = pre_{i-1}(x)+pre_{i}(x)+pre_{i+1}(x),\ i = 3 and 4, 
\end{equation}
where $m_3$ aggregates features captured from the pre-trained model's twice, third, and fourth stages. $m_4$ aggregates features captured from the pre-trained model's third, fourth, and fifth stages.

\begin{equation}
\label{eq6}
 m_{5} = pre_{5}(x)+pre_{6}(x),
\end{equation}
where $m_5$ aggregates features extracted from the fourth and fifth stages of the pre-trained model.

After that, the DANet inputs these multi-level features into multi-scale adaptive fusion modules to selectively fuse features and develops deep features with sizes of $64 \times 64$, $32 \times32$, $16\times16$, and $8 \times 8$, as Equation \ref{eq7}.

\begin{equation}
\label{eq7}
 a_{i} = F_{MAFM}(m_{i}), 
\end{equation}
where $F_{MAFM}(\cdot)$ indicates the multi-scale adaptive fusion module, and $i=2,3,4,5$. 

DANet fuses top-down multi-level features with fine details through  multi-scale channel spatial aggregate modules (CRM), as Equation \ref{eq8}.
 \begin{equation}
\label{eq8}
 s_{i} = F_{CRM}(a_{i}), 
\end{equation}
where $F_{CRM}(\cdot)$ indicates the multi-scale channel spatial aggregate module, and $i=2,3,4,5$. 

Finally, DANet upsamples predictions through deconvolution layer, as Equation \ref{eq9}.
 \begin{equation}
\label{eq9}
 d_{i-1} = s_{i-1}+Dconv2d(s_{i}), 
\end{equation}
where $Dconv2d(\cdot)$ represents the deconvolution operation, and $i=2,3,4,5$.

\subsection{Multi-Scale Adaptively Fusion Module}

\begin{figure*}[htbp]
    \centering
 \includegraphics[width=18cm]{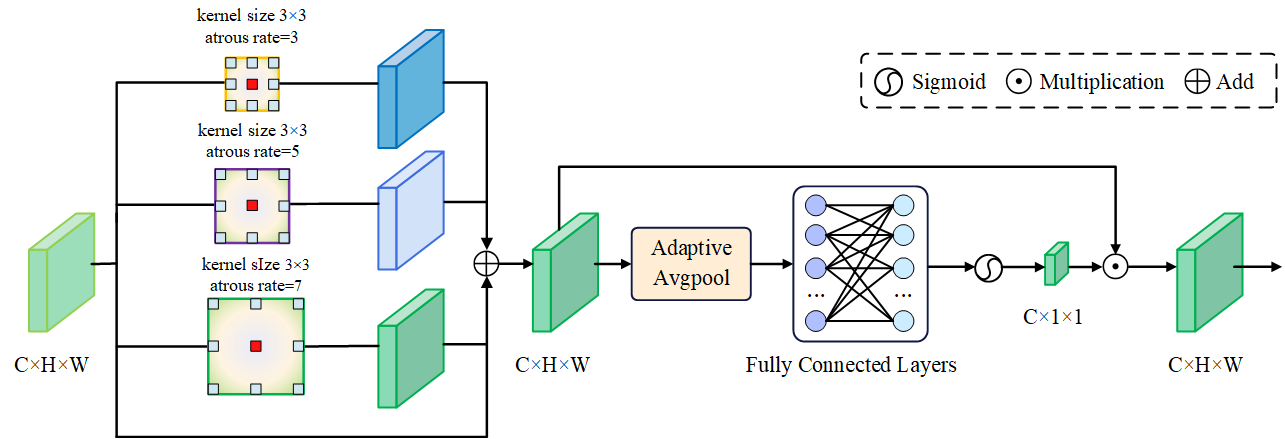}
	\caption{The overview of the multi-scale adaptive fusion module (MAFM).\label{fig3}}
\end{figure*}

The DANet utilizes multi-scale adaptive fusion modules (MAFM) to automatically focus on object structures of multiple scales and implicitly suppress irrelevant features. As shown in figure \ref{fig3}, the MAFM module inputs features integrated by the nearby stages of the pre-trained model and extracts multi-scale features through parallel atrous convolution with different rates, as Equation \ref{eq101}.
\begin{equation}
\label{eq101}
mid\_a_{i} = Atrous_{3}(a_{i})+Atrous_{5}(a_{i})+Atrous_{7}(a_{i}), 
\end{equation}
where $Atrous_{k}(\cdot)$ represents the atrous convolution with $3\times3$ kernel size, $i=2,3,4,5$ and $k$ atrous rate.

Then, the MAFM module fuses muli-scale features and initial features through the residual connection and utilizes adaptive max pooling to obtain weights for each channel. After that, the MAFM module uses fully connected layers and a sigmoid function to compress global features into a one-dimensional vector. Moreover, the MAFM restores feature size by multiplying the weights of each channel and initial features, as Equation \ref{eq111}.

\begin{equation}
\begin{split}
    \label{eq111}
F_{MAFM}(m_{i})=a_{i} + mid\_a_{i} \otimes F_{Sigmoid}(Conv2d\\(Conv2d(F_{Adaptive\ Avgpool}(m_{i})))), 
\end{split}
\end{equation}
where $\otimes$ denotes the dot-multiply operation, $Conv2d(\cdot)$ represents the convolution with a ReLU function and a batch normalization, and $F_{Adaptive\ Avgpool}(\cdot)$ represents the adaptive average pooling operation.

\vspace{0.1cm}
\subsection{Context Refinement Module} 

\begin{figure}[htbp]
    \centering
 \includegraphics[width=7cm]{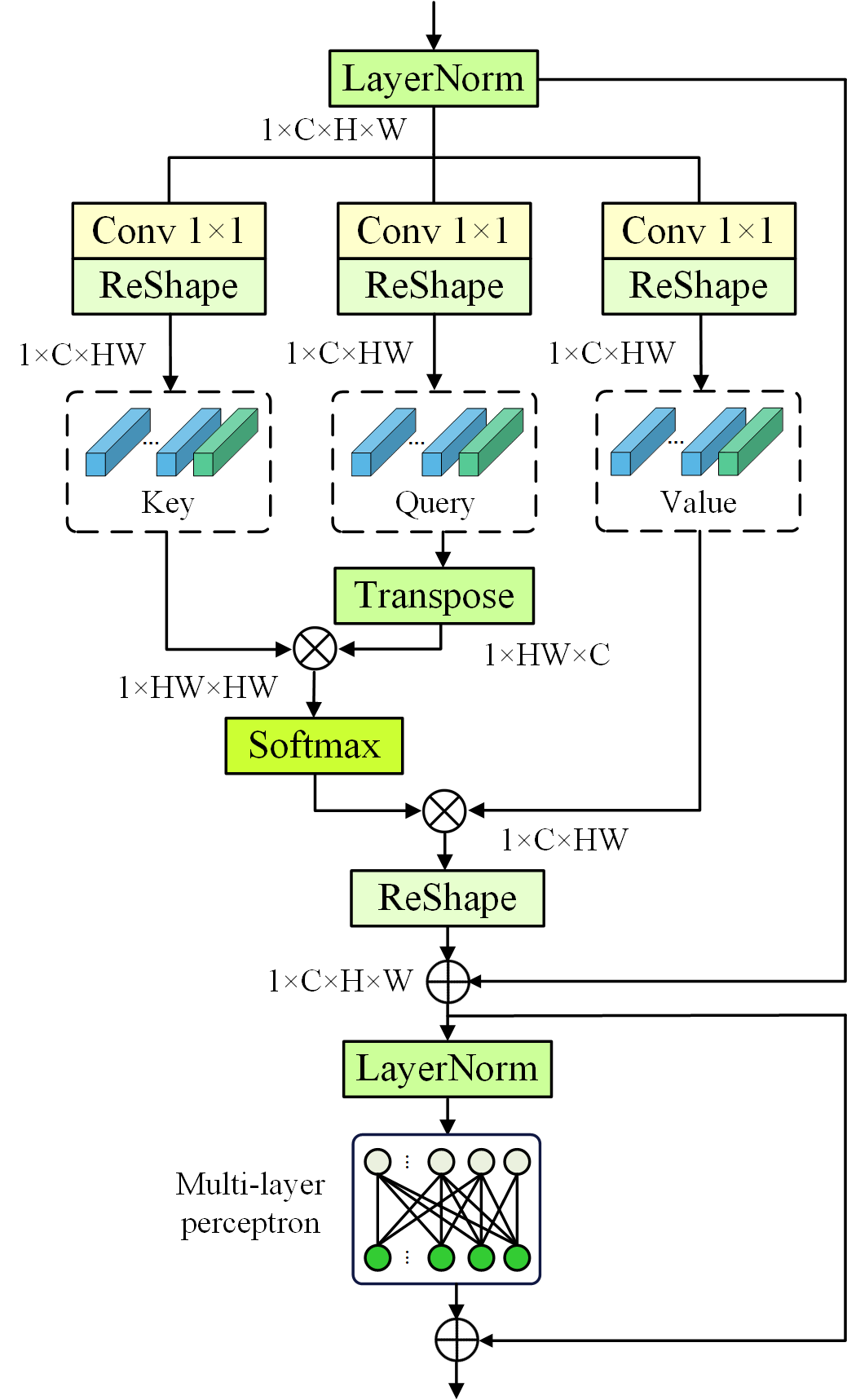}
	\caption{The overview of the context refinement module (CRM).\label{fig4}}
\end{figure}

As shown in figure \ref{fig4}, the DAGAN framework uses context refinement modules (CRM) to transform more exhaustive contextual information into local features, which could assist the framework in constructing contextual dependencies over global features and enhance the representation capability. Next, we elaborate on the processing of the CRM module.

We represent the convolution operation as Equation \ref{eq05}, where ${W}^{{n} \times {n}}$ indicates the ${n} \times {n}$ convolution kernel, ${b}$ denotes the bias vector.
\begin{equation}\label{eq05}
W^{n \times n}(x)=W^{n \times n} \odot x+b
\end{equation}

In the CRM module, the features $x \in R^{C \times H \times W}$ are input fed into the layer normalization and convolution operation to generate new features $Key^1$, $Query^1$ and $Value^1 \in R^{C \times H \times W})$, which are defined as Equation \ref{eq06}.

\begin{align} \label{eq06}
\left\{\begin{matrix}
Key^1 =W_1^{1 \times 1} \odot x+b_1 \\
Query^1 =W_2^{1 \times 1} \odot x+b_2 \\
Value^1 =W_3^{1 \times 1} \odot x+b_3
\end{matrix}\right.    
\end{align}

Then, $Key^1$, $Query^1$ and $Value^1$ are reshaped to $Key^2$, $Value^2$, $Query^2 \in R^{1 \times C \times H W}$, as Equation \ref{eq07}.

\begin{align}\label{eq07}
\left\{\begin{matrix}
Key^2=F_{Reshape}(Key^1) \\
Query^2=F_{Reshape}(Query^1)\\
Value^2=F_{Reshape}(Value^1)
\end{matrix}\right.
\end{align}

$Query^2 \in R^{1 \times C \times H W}$ is transposed to $ Query^3 \in R^{1 \times HW \times C}$, as Equation \ref{eq08},.
\begin{equation}\label{eq08}
Query^3=F_{Transpose}(Query^2)
\end{equation}

$ Query^3 \in R^{1 \times HW \times C}$ and $Key^2 \in R^{1 \times C \times H W}$ perform matrix multiplication operation $ \otimes$, and compute spatial context features through the softmax layer $F_{softmax}(\cdot)$, as Equation \ref{eq09}.
\begin{equation}\label{eq09}
F_{QM}(x)=F_{softmax}(Query^3 \otimes Key^2)
\end{equation}

After that, the result of the softmax layer $F_{SM}(x) \in R^{1 \times H W \times H W} $ and $Value^2 \in R^{1 \times C \times H W}$ perform matrix multiplication operation, as Equation \ref{eq10}.

\begin{equation}\label{eq10}
F_{VM}(x)=Value^2 \otimes F_{SM}(x)
\end{equation}

And CRM module reshapes $F_{VM}(x)=$ to $ R^{1 \times C \times H \times W}$, inputs them into layer normalization and multi-layer perception $MLP(\cdot)$, as Equation \ref{eq11} shows.
\begin{equation}\label{eq11}
CRM(x)=MLP(F_{Reshape}(F_{VM}(x)\oplus x))
\end{equation}

\subsection{Loss Function} \label{sec3.4}
 
The generator and the discriminator are trained simultaneously: the framework would automatically adjust the generator's parameters so that the image generated by the generator is getting closer to the real image, i.e., minimize $\log (1-D(G(z)))$. And the DAGAN framework adjusts the parameters to make the discriminator more capable of correctly judging whether the data comes from real scenes, i.e., maximizing $\log {D}(X)$. More specifically, the generative loss of the DAGAN framework as Equation \ref{eq02}, and the adversarial loss as Equation \ref{eq02},
\begin{equation}
\begin{aligned}
\label{eq02}
L_{G}(x_1,x_2)=-E_{x_1, x_2, y}(\log (D(G(x_1,x_2)))), 
\end{aligned}
\end{equation}
where  $x_1$ and $x_2$ represent bi-temporal images, and $G(x_1,x_2)$ means the output of the generator,.

Additionally, the adversarial loss is defined as Equation \ref{eq090}. $y$ represents the ground truth, and $D(G(x_1, x_2))$ represents the output of the discriminator. The term $l_{BCE}(\cdot$ and $l_{Dice}(\cdot)$ indicates cross-entropy loss and dice loss, which could alleviate class imbalance problem.

\begin{equation}
\begin{aligned}
\label{eq090}
L_{D}(x_1,x_2,y)=&-E_{x,y}(\log (D(y)) + \log (1-D(G(x_1,x_2)))) \\
+ & \sum_{i=1}^{n}L_{BCE}(y, G(x_1,x_2)) \\
+ &\sum_{i=1}^{n}L_{Dice}(y, G(x_1,x_2)), 
\end{aligned}
\end{equation}
where $i$ indicates the $i-th$ level of the decoder, and $n=4$.

\section{Experiments}
\subsection{DataSet}

\begin{figure}[H]
    \centering
        {\includegraphics[width=.3\linewidth]{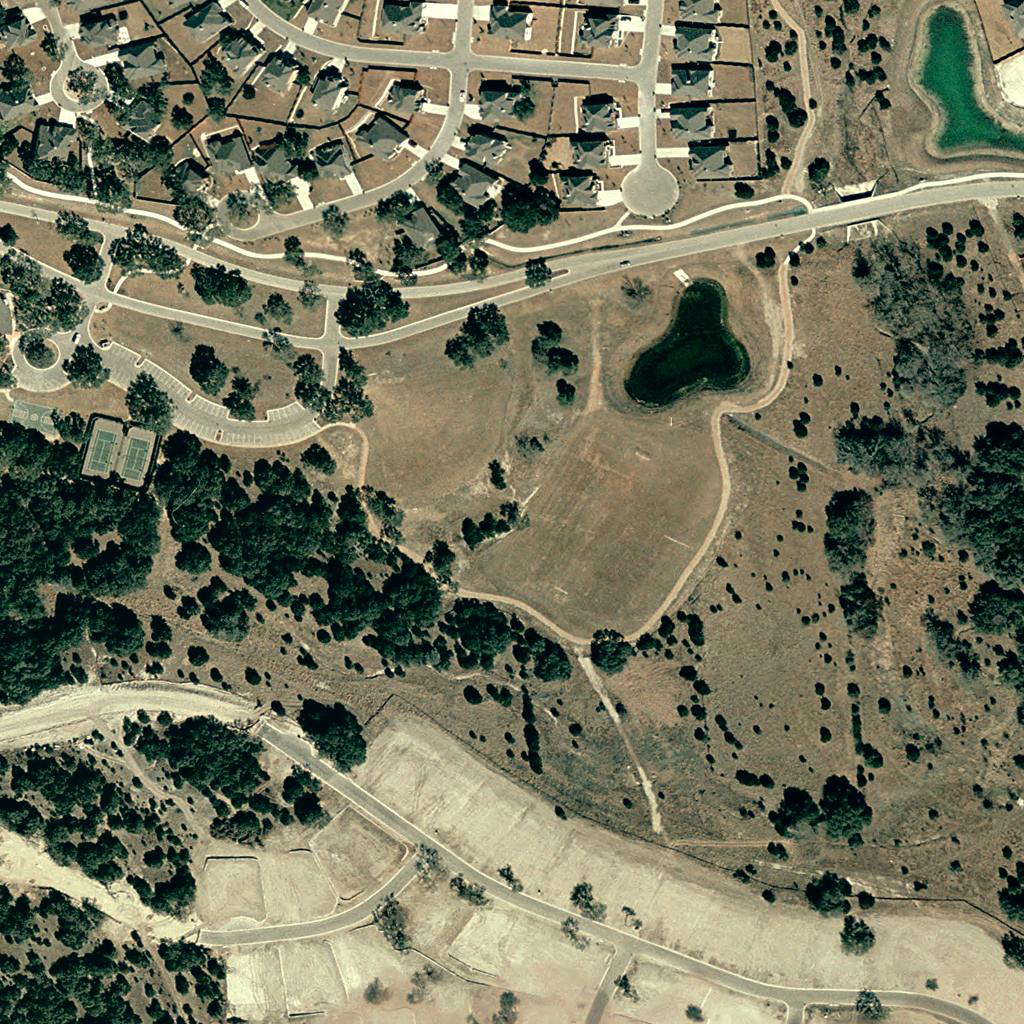}} 
	  {\includegraphics[width=.3\linewidth]{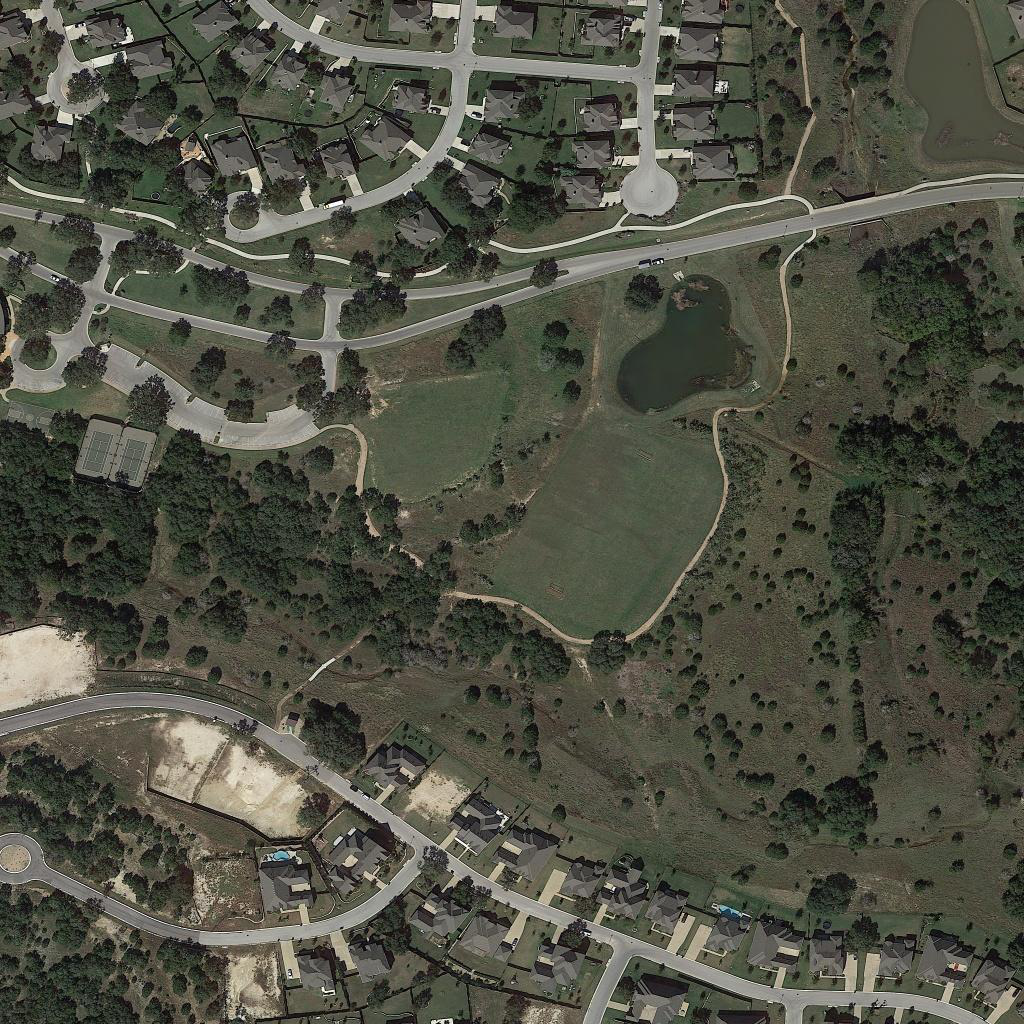}}
        {\includegraphics[width=.3\linewidth]{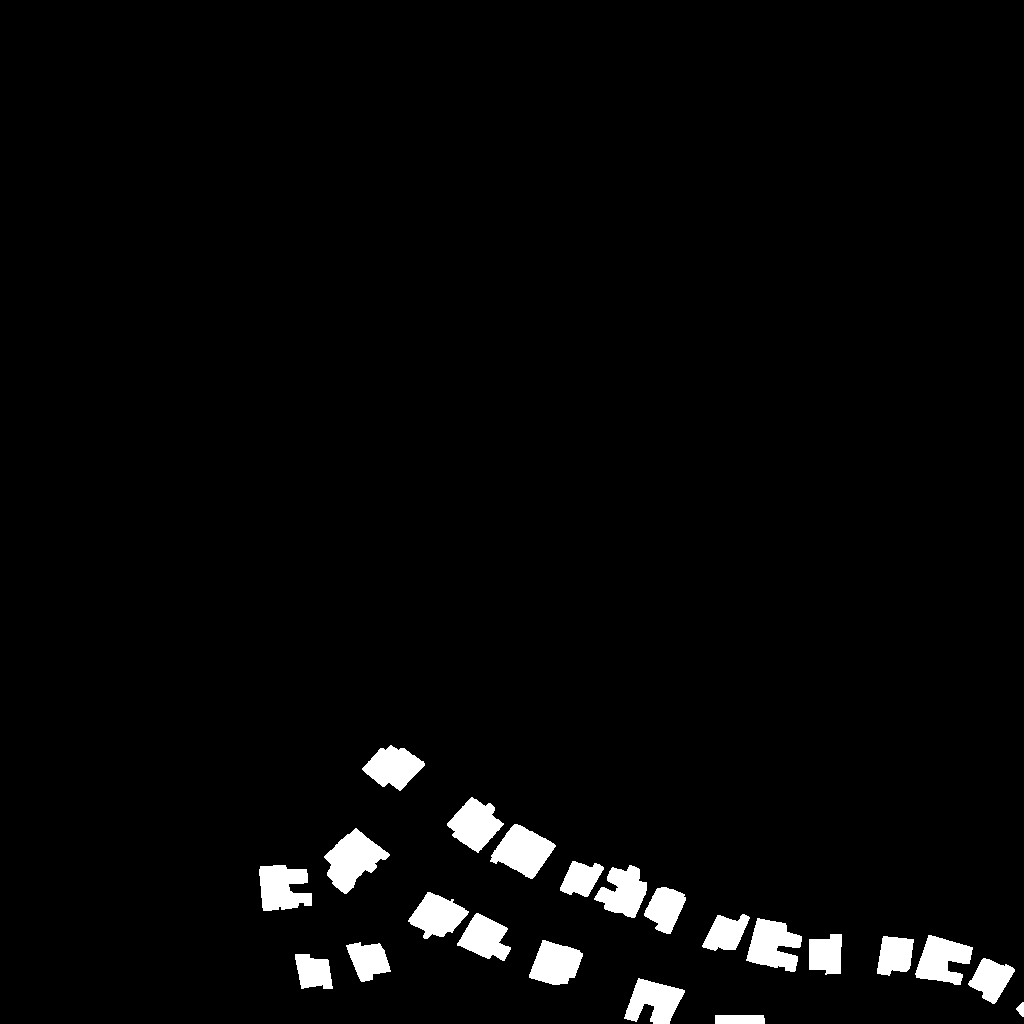} }
        
        \subfigure[T1 images]{\includegraphics[width=.3\linewidth]{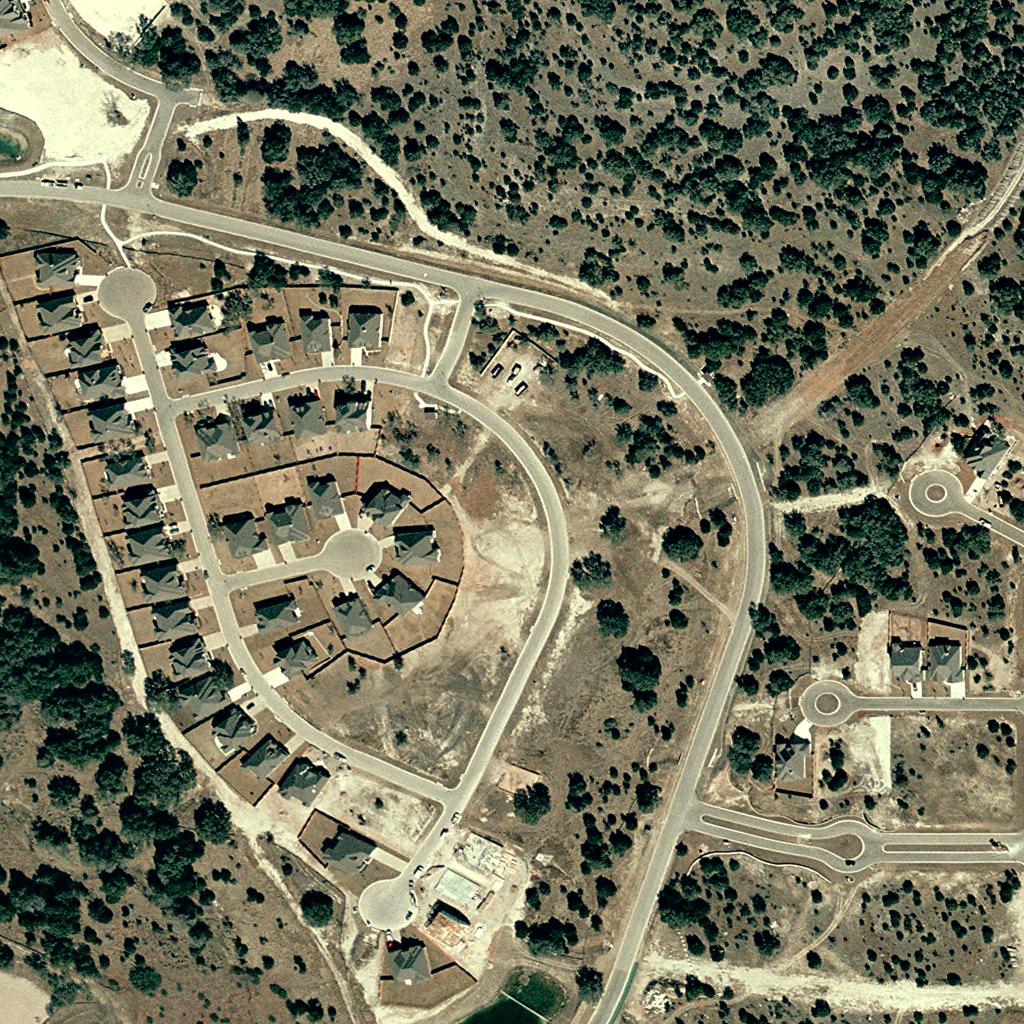}} 
	  \subfigure[T2 images]{\includegraphics[width=.3\linewidth]{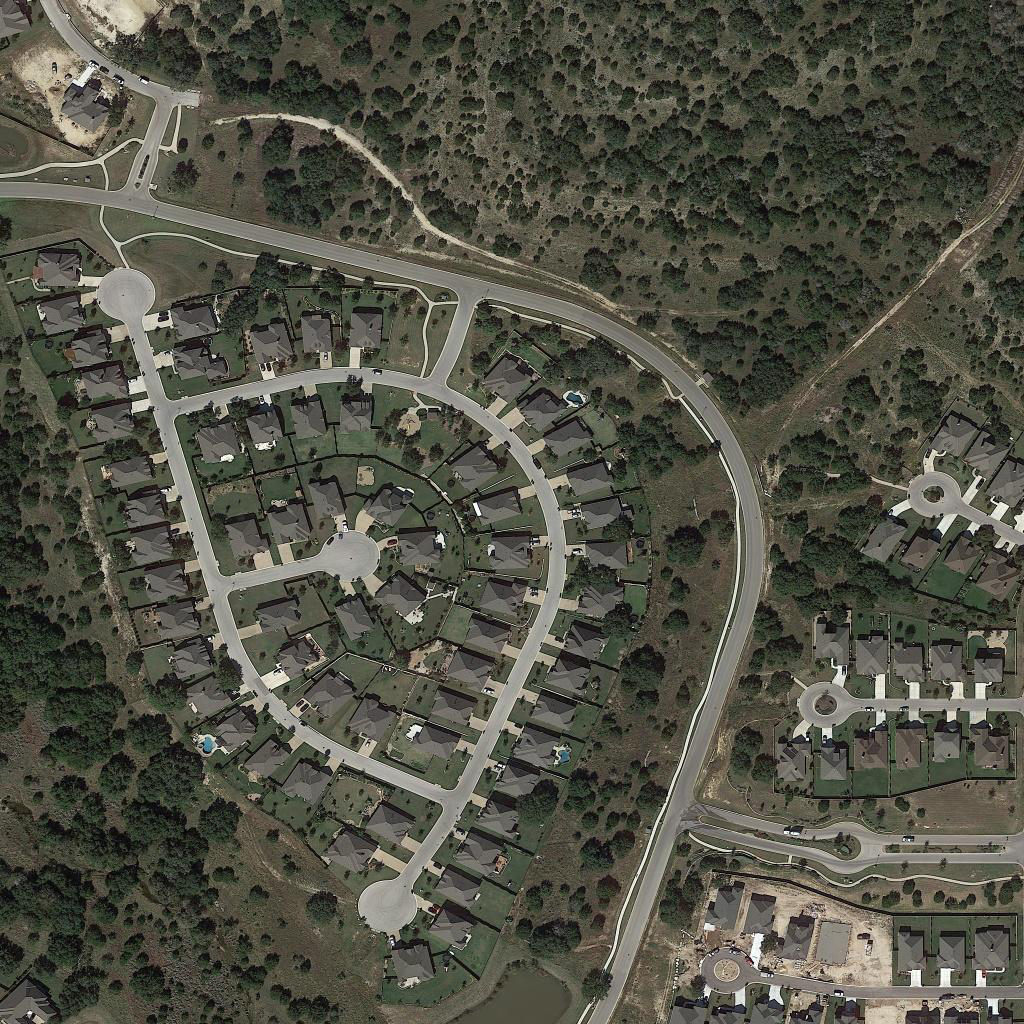}} 
        \subfigure[Ground truth]{\includegraphics[width=.3\linewidth]{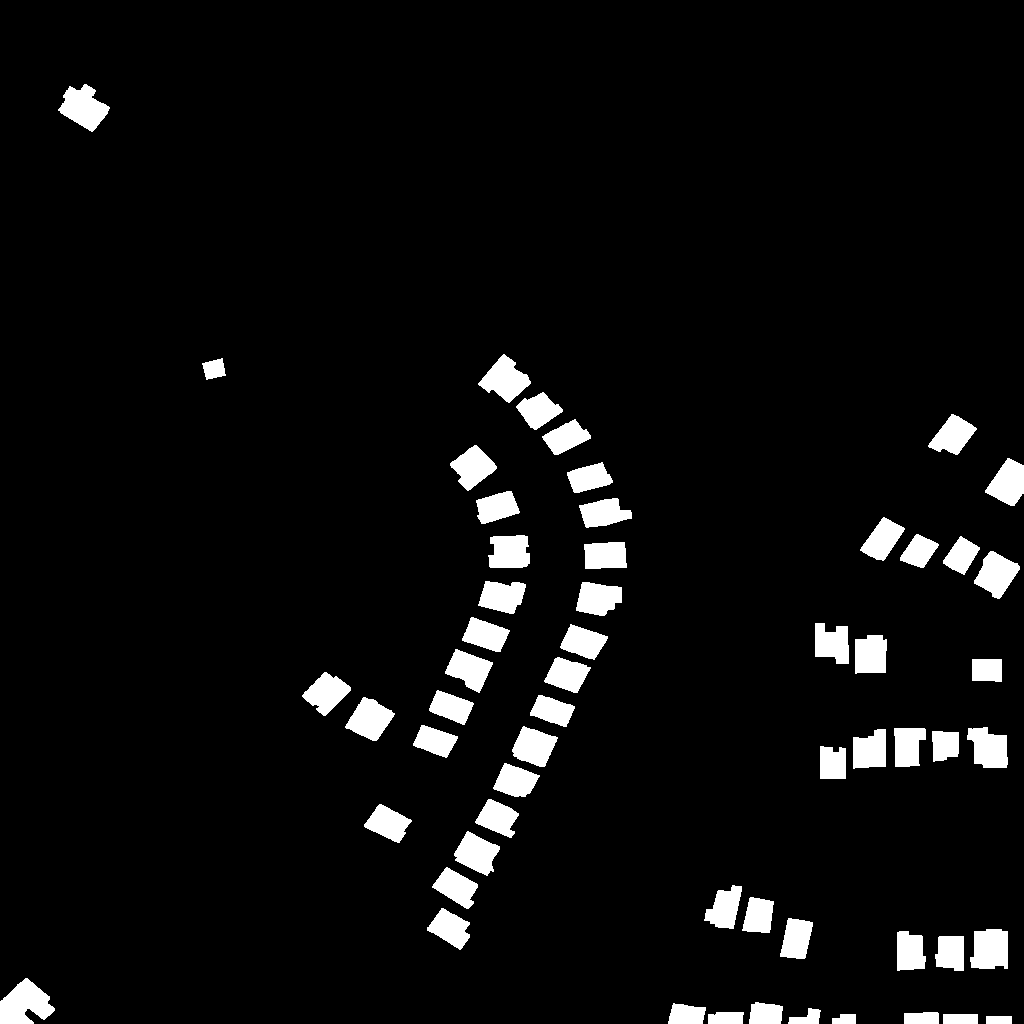} }
	\caption{Some samples of the LEVIR datasets.\label{fig6}}
\end{figure}

Detecting the change of VHR images is a significant task in various applications. The LEVIR dataset has 637 Google Earth image pairs with $1024 \times 1024$ pixels, which were collected from 20 regions in the US. We adopt the official dataset setting (7:1:2) as the experimental dataset to divide images into patches with $256 \times 256$ pixels. In experiments, 7120 paired images are used as the training set, 1024 and 2048 paired images are utilised as the validation and test sets. Figure \ref{fig6} shows several samples of the dataset. 

\subsection{Evaluation Metrics}
We choose $precision$, $recall$, $f1 score$, $kappa$ and $Intersection over Union (IoU)$ to evaluate the performance of the DAGAN framework and other advanced methods, where $TP$, $TN$, $FP$, and $FN$ indicate true positive, true negative, false positive, and false negative, respectively.
\begin{equation}
	Precision=\frac{T P}{T P+F P}
\end{equation}

The $recall$ represents the frequency of correct pixels among the instances labeled as positive pixels.
\begin{equation}
	Recall =\frac{T P}{T P+F N}
\end{equation}

The $F1$ is estimated by precision and recall, which could assess the model's effectiveness.
\begin{equation}
	{ F1\; score }=2 \times \frac{ { Precision } \times  { Recall }}{ { Precision }+ { Recall }}
\end{equation}

The $OA$ is the proportion of correct pixels to the total pixels.
\begin{equation}
	{ OA }=\frac{T P+T N}{T P+T N+F P+F N}
\end{equation}

The $kappa$ coefficient represents generating errors of completely random classification of the proportion.
\begin{equation}
P_{e}=\frac{(T_{n}+F_{n})\cdot (T_{n}+F_{p})+(F_{p}+T_{p})\cdot (F_{n}+T_{p})}{(T_{p}+F_{p}+T_{n}+F_{n})^2} 
\end{equation}
\begin{equation}
kappa=\frac{OA-P_{e}}{1-P_{e}} 
\end{equation}

The $IoU$ calculates the region similarity between the prediction generated by the framework and the ground truth. 
\begin{equation} \label{equ9}
    IoU=\frac{X\cap Y}{X\cup Y}
\end{equation}
where $X\cap Y$ denotes the overlapping regions of $X$ and $Y$, and $X\cup Y$ indicates the region $X$ and region $Y$.

\subsection{Implement Details}

In training, we utilize the ResNet-50 pre-trained models on ImageNet to initialize the DAGAN framework. Then, We use an Adam optimizer with 0.9 momentum and 0.0001 weight decay, in which $\beta_1$ and $\beta_2$ are set as 0.9 and 0.99, respectively. The learning rate is set as 0.0005 and reduces as $(1-(cur_-iteration/max_-80000))^{0.9}\times lr$. We set the iteration to 80000 and the batch size to 16. We adopt data augmentation to increase the training set, such as random flipping, cropping, etc. 



\subsection{Ablation Experiments}

\begin{table*}[htb]
	\centering

	\caption{Ablation experiments on the LEVIR test set. 
		\label{tab1}}
	\begin{tabular}{@{}cccccccccccc@{}}
		\toprule
		\multirow{2}{*}{Method} & \multirow{2}{*}{ResNet50} & \multirow{2}{*}{Aggregate Connection} & \multirow{2}{*}{MAFM} & \multirow{2}{*}{CRM} & \multirow{2}{*}{GAN} & \multirow{2}{*}{$Kappa$} & \multirow{2}{*}{$IoU$} & \multirow{2}{*}{$F1$}  & \multirow{2}{*}{$Recall$} & \multirow{2}{*}{$Prec$} \\
	& 	&  &                                   &                              &                        &                      &                                  &                       &                        &                      \\
		\midrule
       
       DAGAN-R &  	\checkmark  	& 
		
		&                                     &                              &                        &           88.17\%           &        79.71\%                                       &     88.71\%         &   87.74\%   & 89.70\%    \\
		DAGAN-A &  \checkmark   & 	\checkmark                         &                               &             &              &  89.61\%                  &  81.97\%                                          &  90.09\%         &  90.97\%  &    89.23\%   \\
  
	DAGAN-M &\checkmark & \checkmark 	& \checkmark                                 &                              &                        &   90.35\%                & 83.15\%                                          &  90.80\%         &  91.76\% &  89.86\%  \\
	DAGAN-MC &  \checkmark   	&              \checkmark                       &    \checkmark    &\checkmark                            &                        &      90.96\%    &       84.75\%                                     &  91.33\%     &    91.23\%    &         91.42\%    \\
   DAGAN &  \checkmark   	&              \checkmark                       &    \checkmark    &\checkmark                            &      \checkmark                     &                 \textbf{ 91.12\% }   &      \textbf{ 85.01\%}                                     &  \textbf{91.48\%}     &    \textbf{91.26\%}    &        \textbf{ 91.69\%}   \\
     
  \bottomrule
	\end{tabular}
\end{table*}

\begin{figure*}[htbp]
	\centering
 
	\begin{minipage}{\linewidth}
		\centering
		{\includegraphics[width=.11\linewidth]{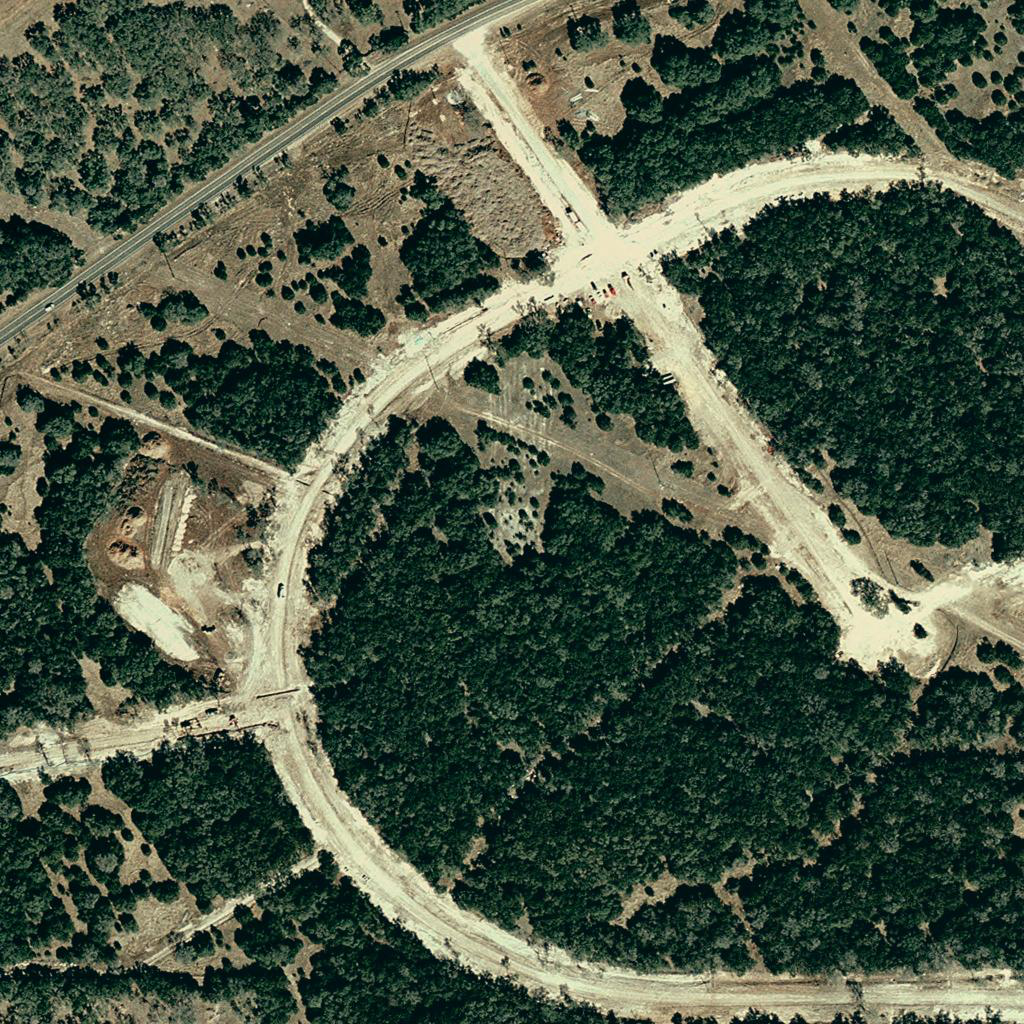}}
		{\includegraphics[width=.11\linewidth]{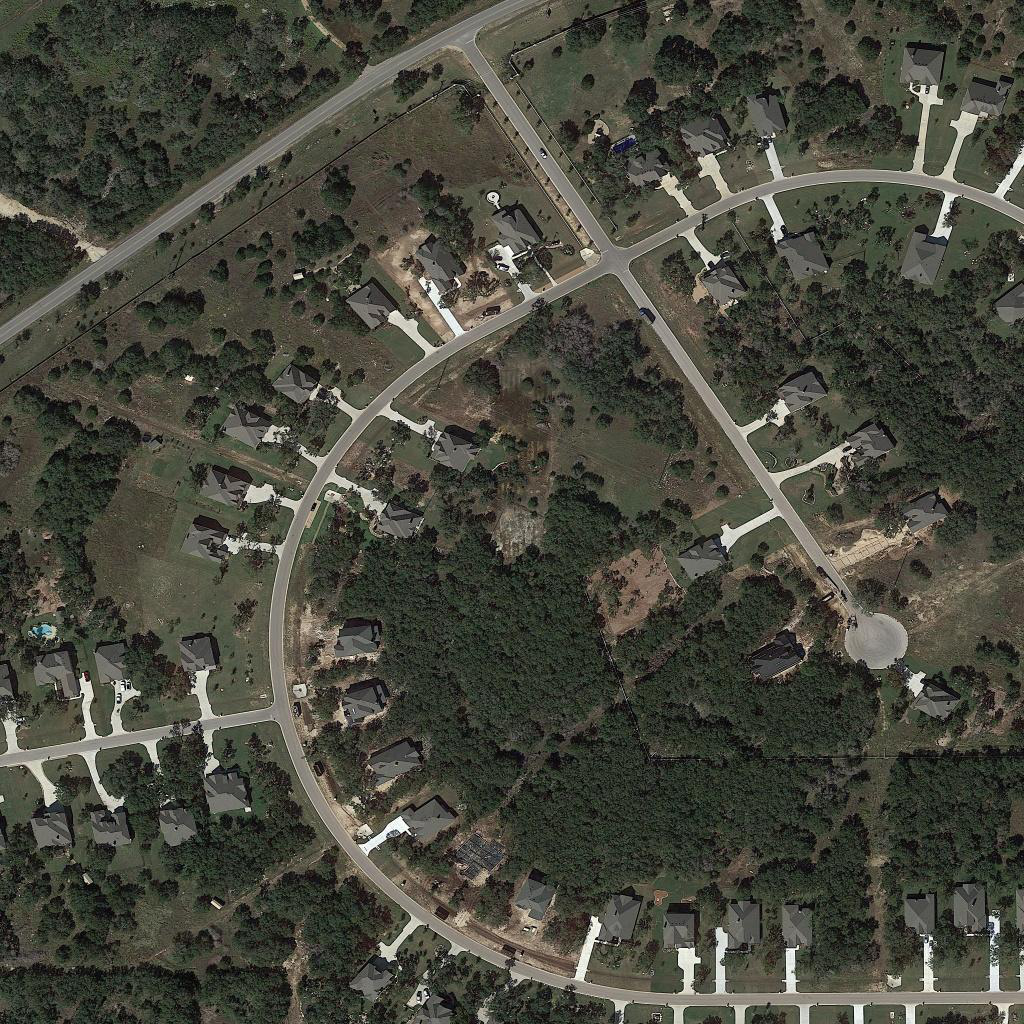}}
		{\includegraphics[width=.11\linewidth]{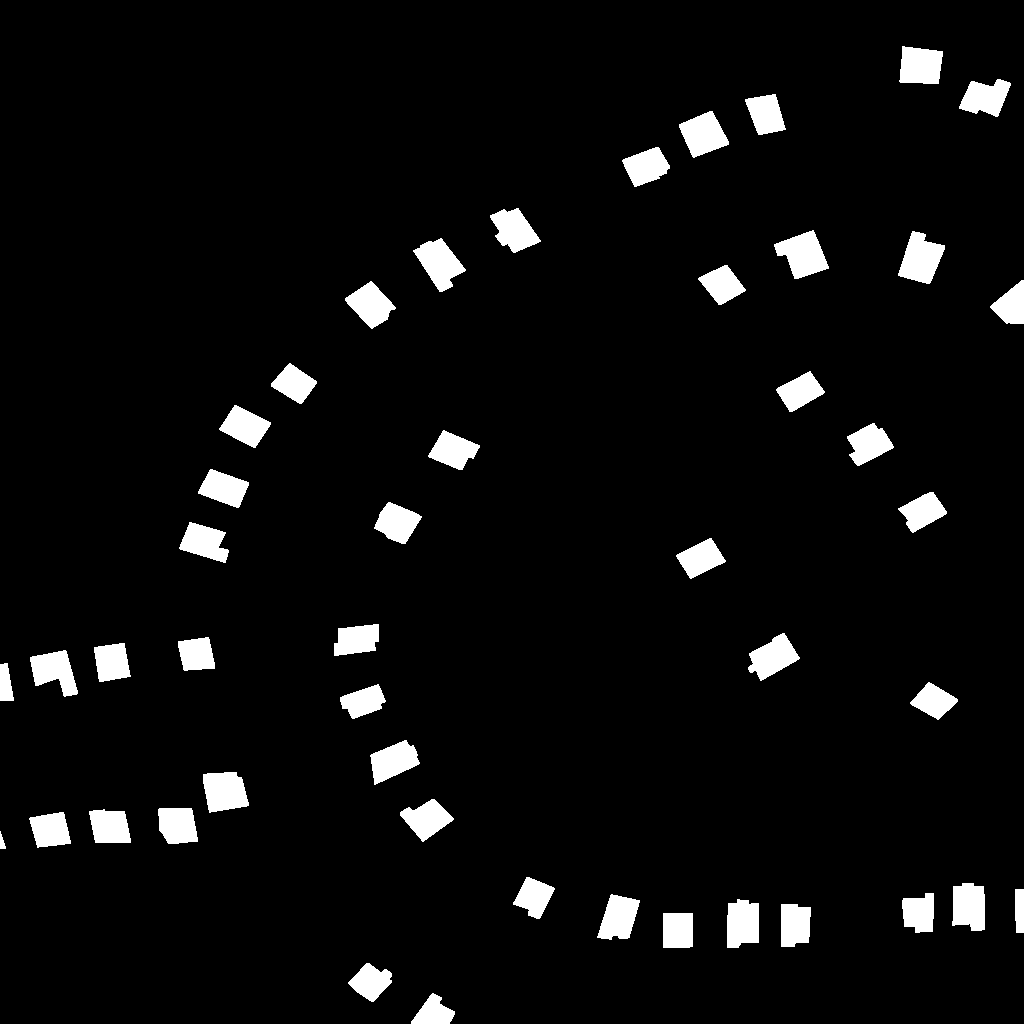}}
		{\includegraphics[width=.11\linewidth]{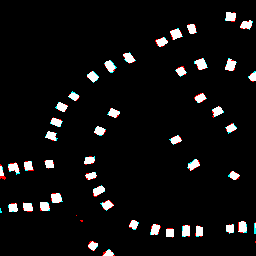}}
		{\includegraphics[width=.11\linewidth]{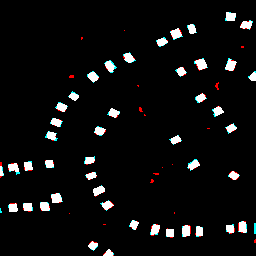}}
		{\includegraphics[width=.11\linewidth]{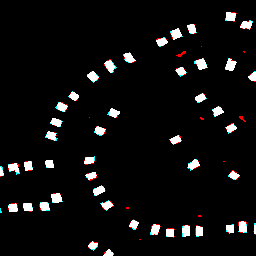}}
		{\includegraphics[width=.11\linewidth]{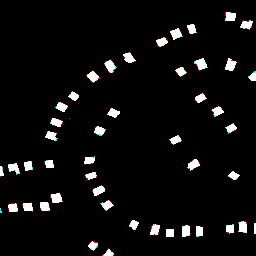}}
		{\includegraphics[width=.11\linewidth]{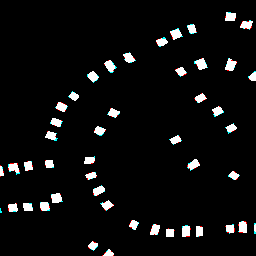}}
	\end{minipage}
	
	\vspace{0.1cm}

	\begin{minipage}{\linewidth}
		\centering
		{\includegraphics[width=.11\linewidth]{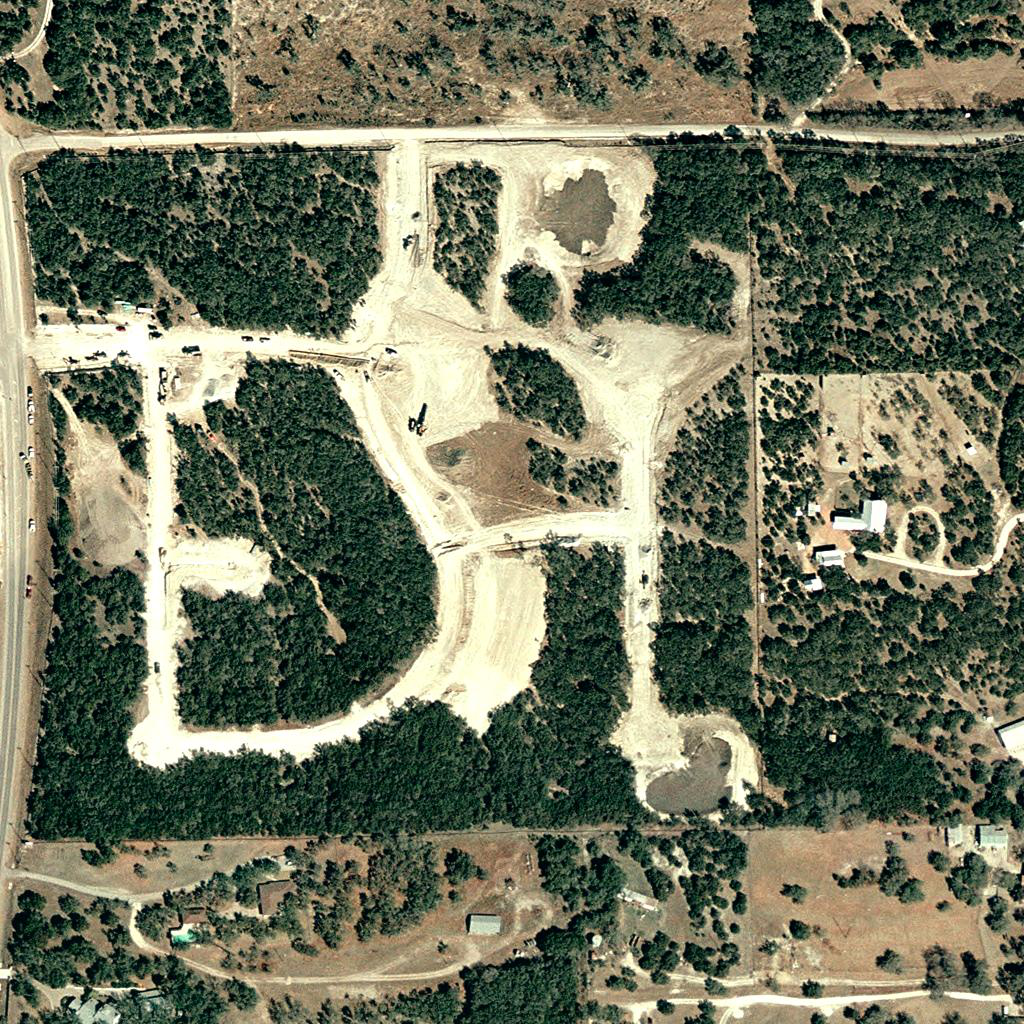}}
		{\includegraphics[width=.11\linewidth]{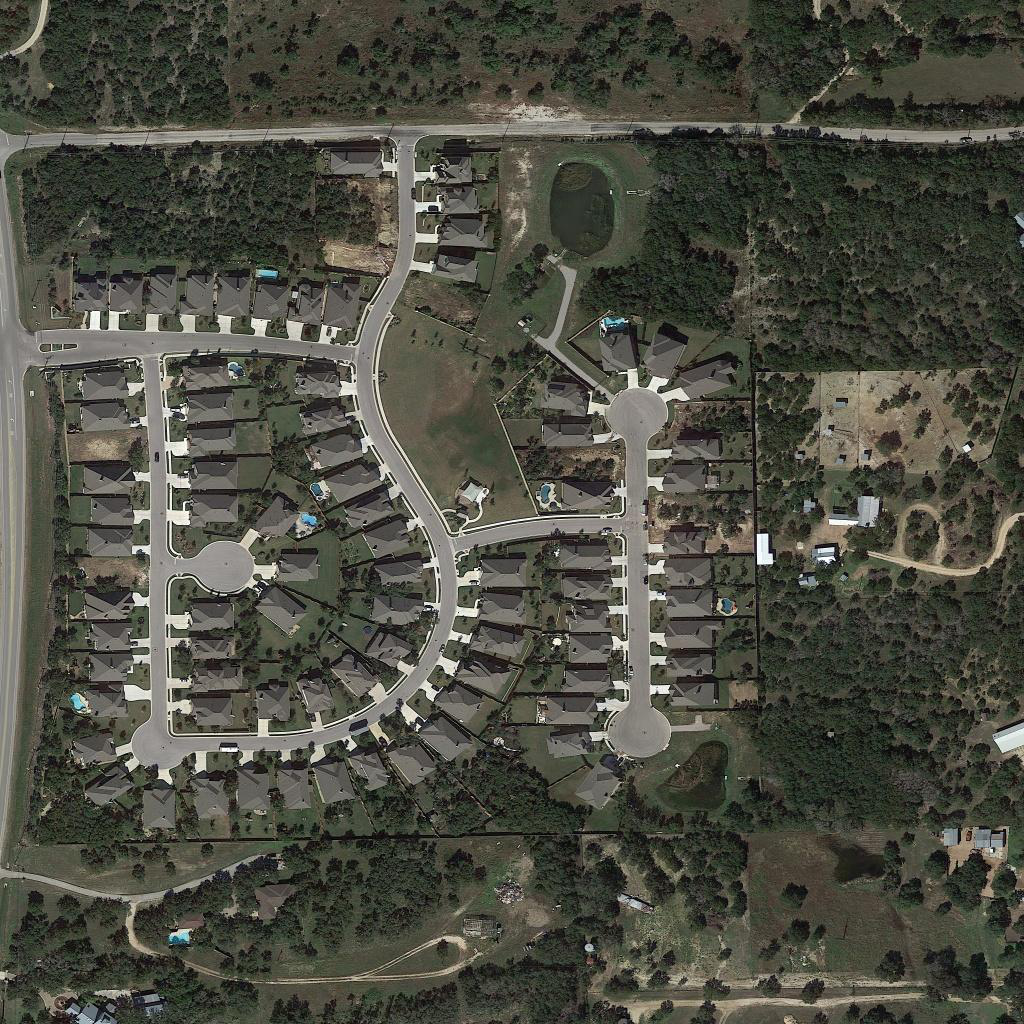}}
		{\includegraphics[width=.11\linewidth]{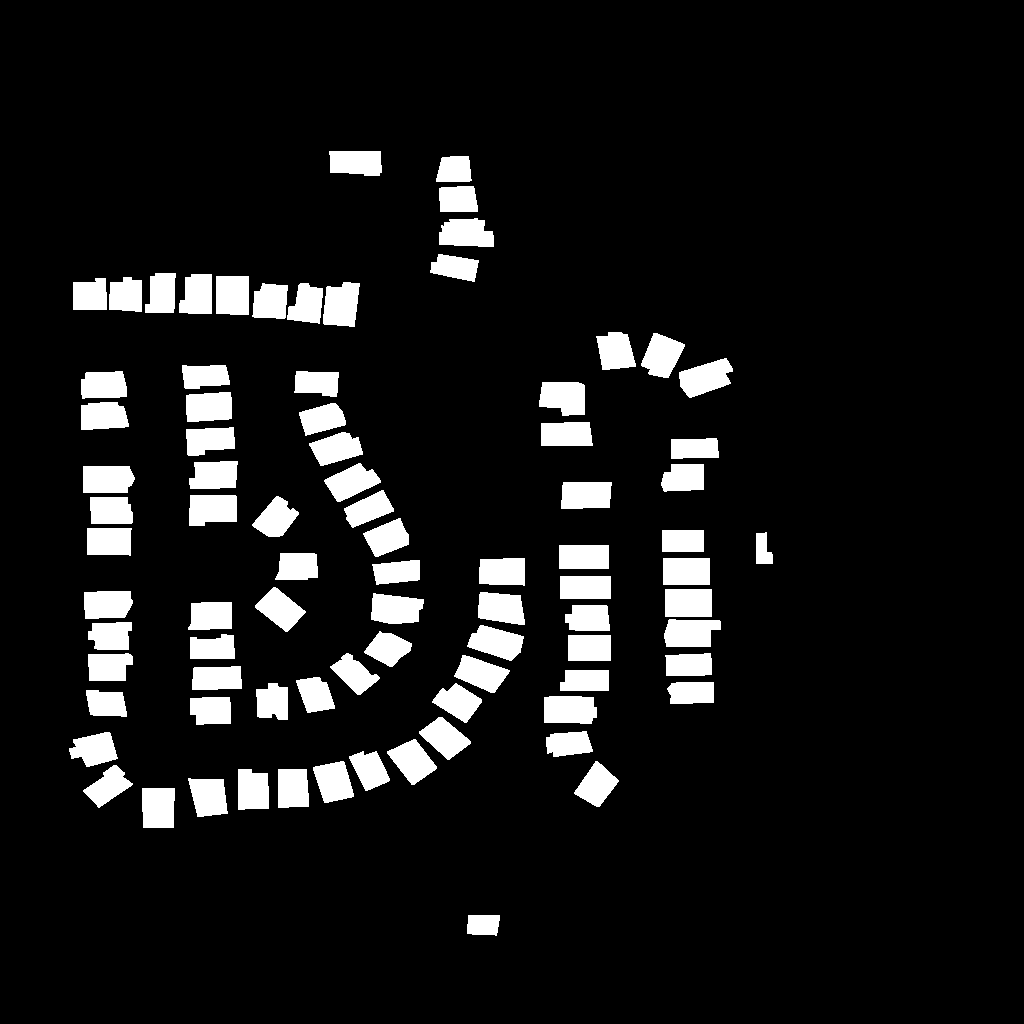}}
		{\includegraphics[width=.11\linewidth]{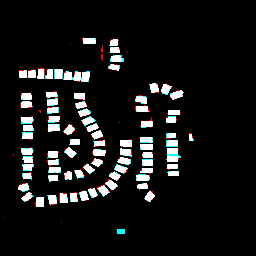}}
		{\includegraphics[width=.11\linewidth]{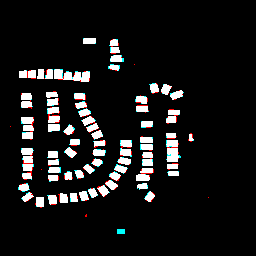}}
		{\includegraphics[width=.11\linewidth]{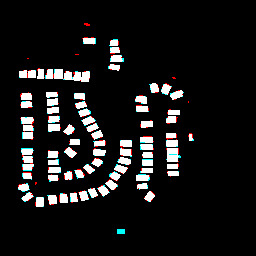}}
		{\includegraphics[width=.11\linewidth]{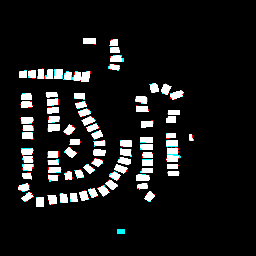}}
		{\includegraphics[width=.11\linewidth]{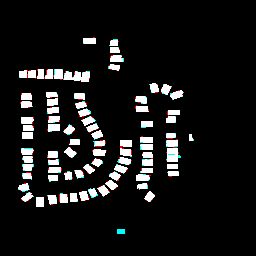}}
	\end{minipage}
	
	\vspace{0.1cm}

	\begin{minipage}{\linewidth}
		\centering
		{\includegraphics[width=.11\linewidth]{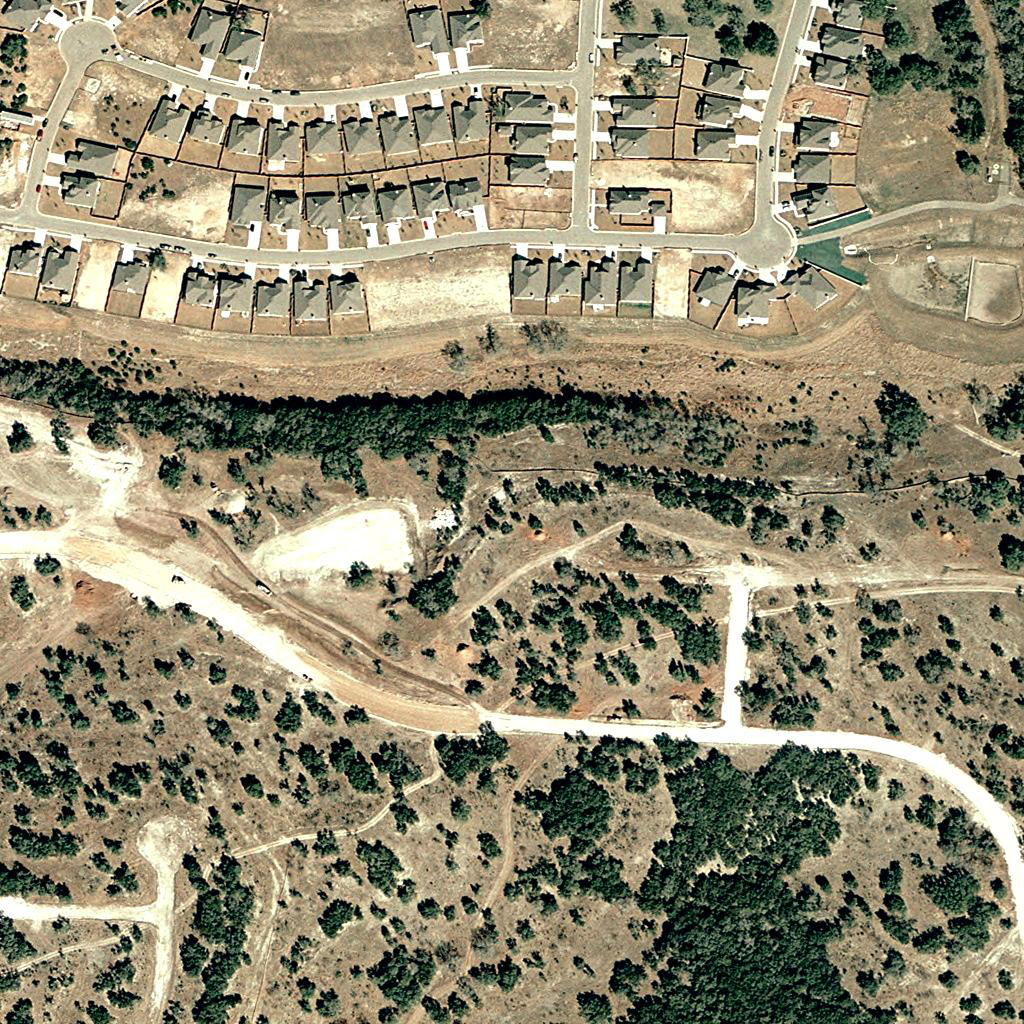}}
		{\includegraphics[width=.11\linewidth]{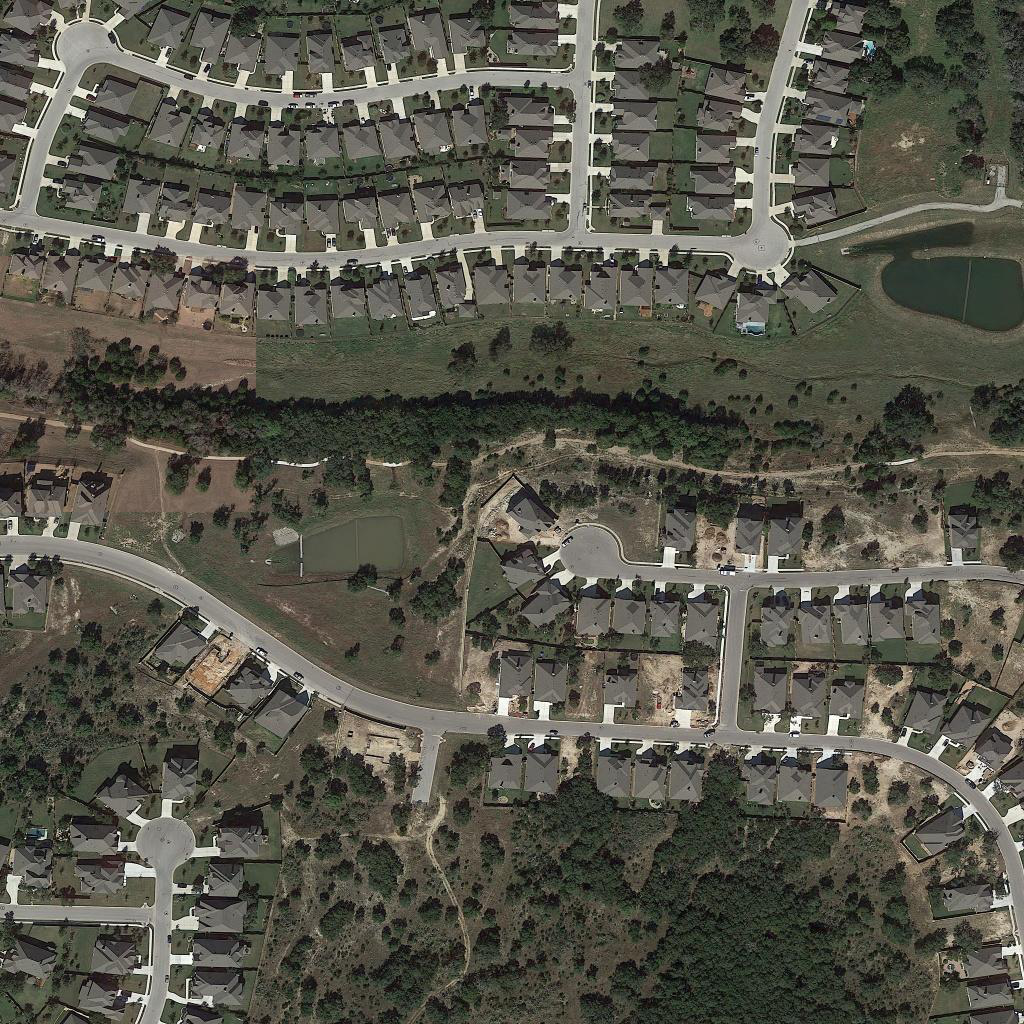}}
		{\includegraphics[width=.11\linewidth]{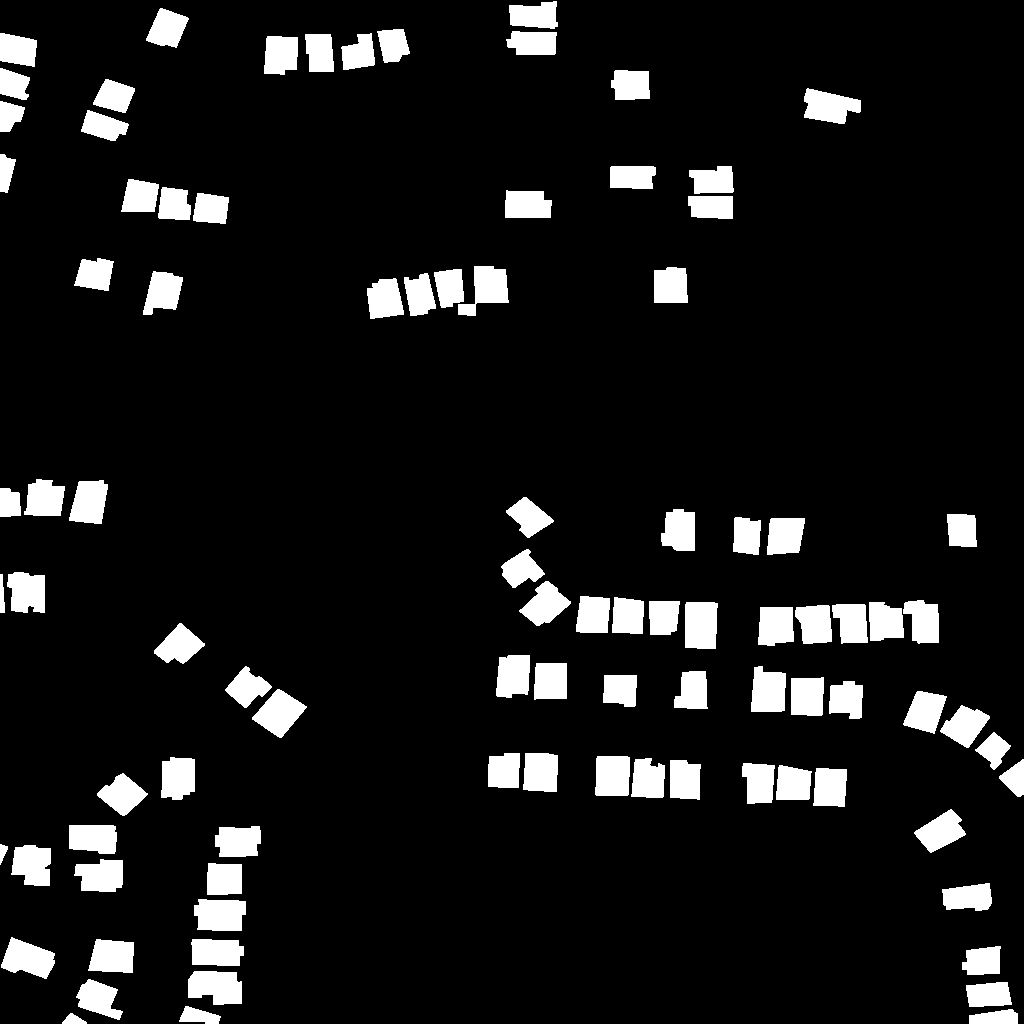}}
		{\includegraphics[width=.11\linewidth]{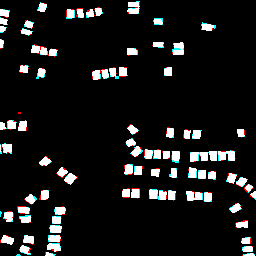}}
		{\includegraphics[width=.11\linewidth]{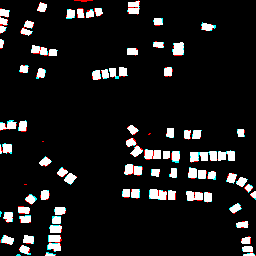}}
		{\includegraphics[width=.11\linewidth]{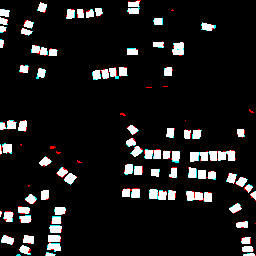}}
		{\includegraphics[width=.11\linewidth]{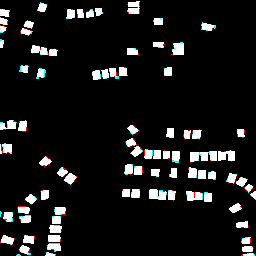}}
		{\includegraphics[width=.11\linewidth]{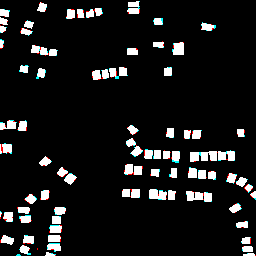}}
	\end{minipage}
	
	\vspace{0.1cm}

	\begin{minipage}{\linewidth}
		\centering
		{\includegraphics[width=.11\linewidth]{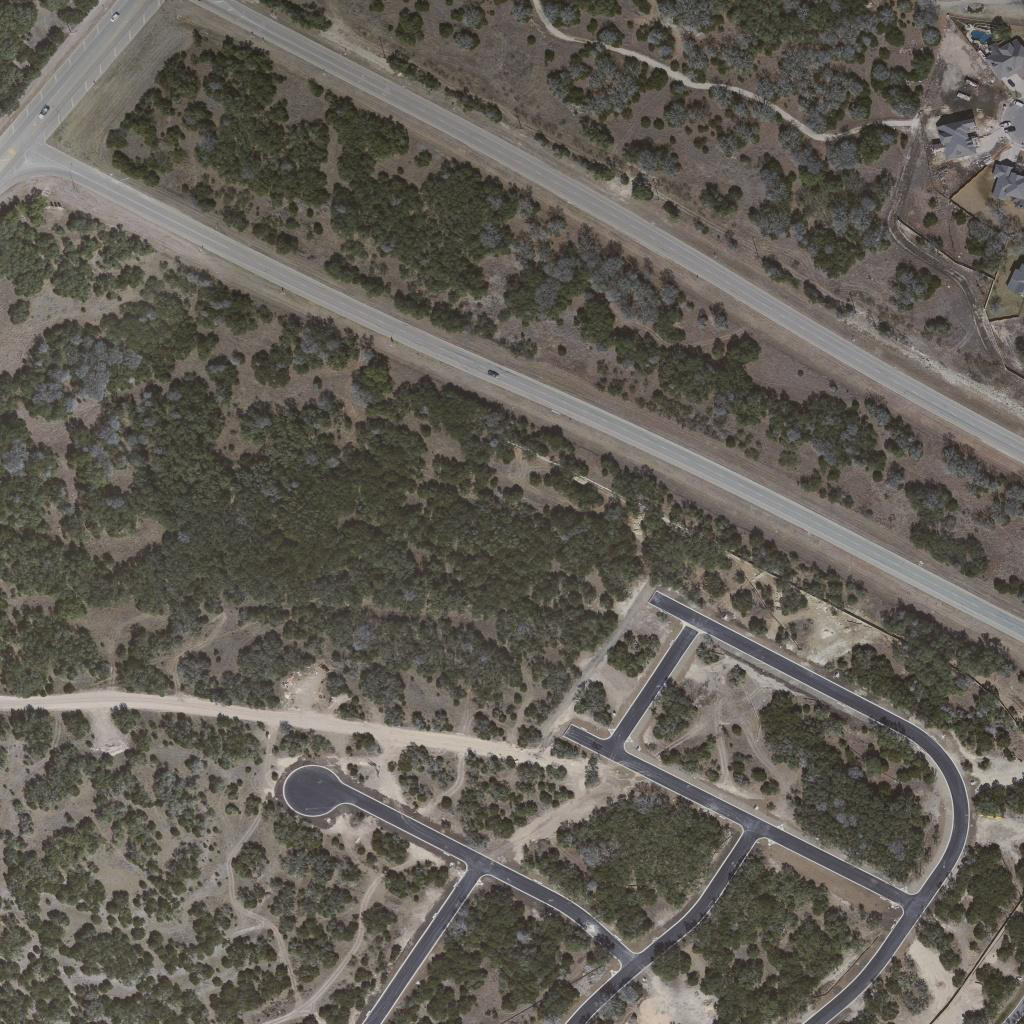}}
		{\includegraphics[width=.11\linewidth]{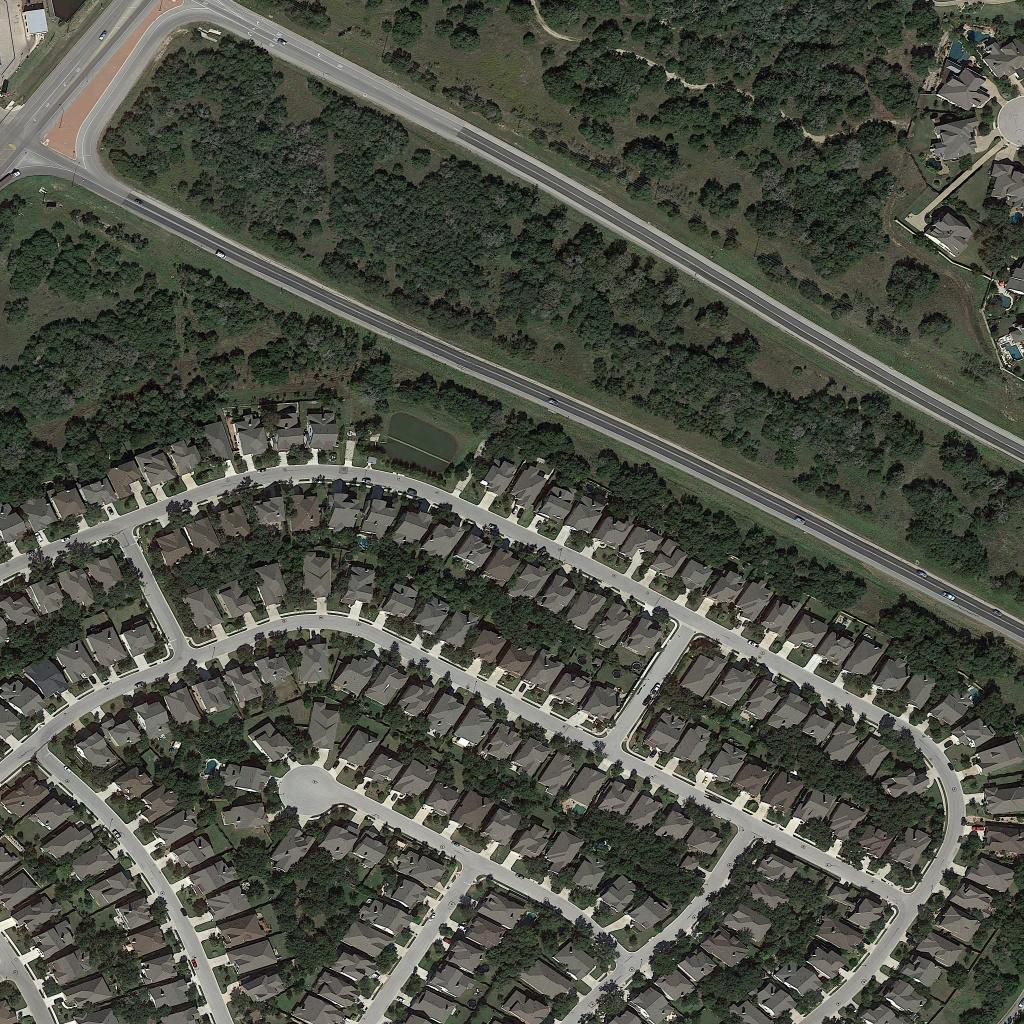}}
		{\includegraphics[width=.11\linewidth]{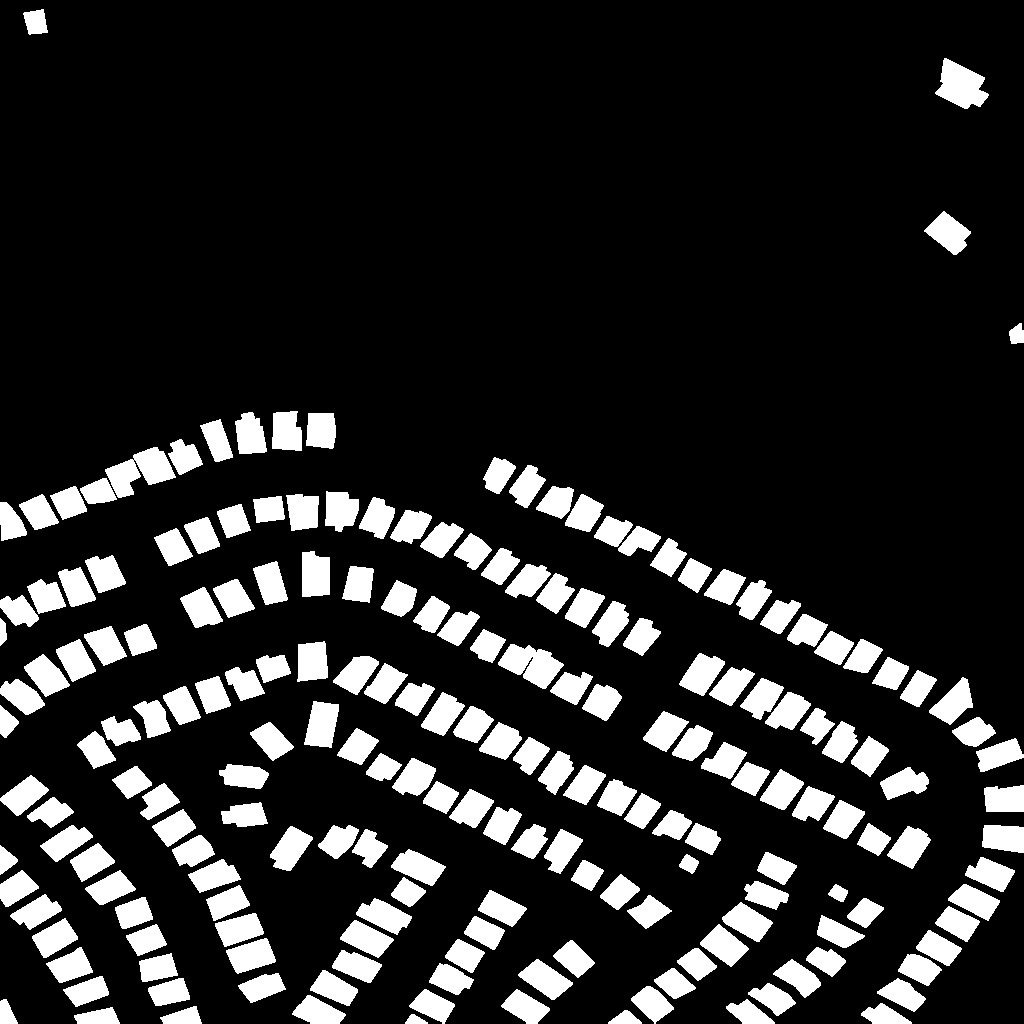}}
		{\includegraphics[width=.11\linewidth]{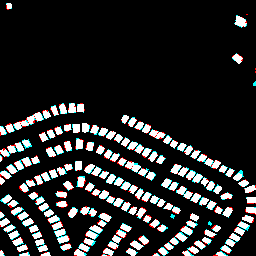}}
		{\includegraphics[width=.11\linewidth]{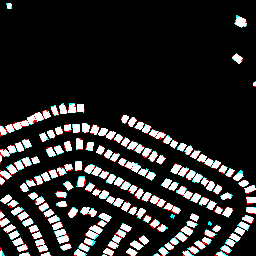}}
		{\includegraphics[width=.11\linewidth]{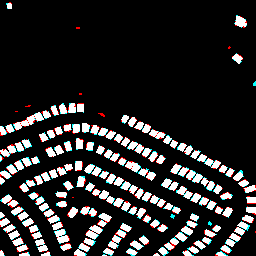}}
		{\includegraphics[width=.11\linewidth]{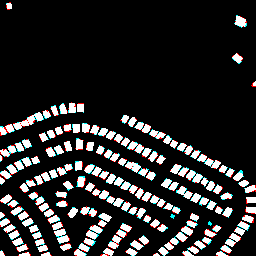}}
		{\includegraphics[width=.11\linewidth]{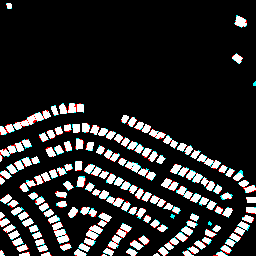}}
	\end{minipage}
	
	\vspace{0.1cm}

	\begin{minipage}{\linewidth}
		\centering
		{\includegraphics[width=.11\linewidth]{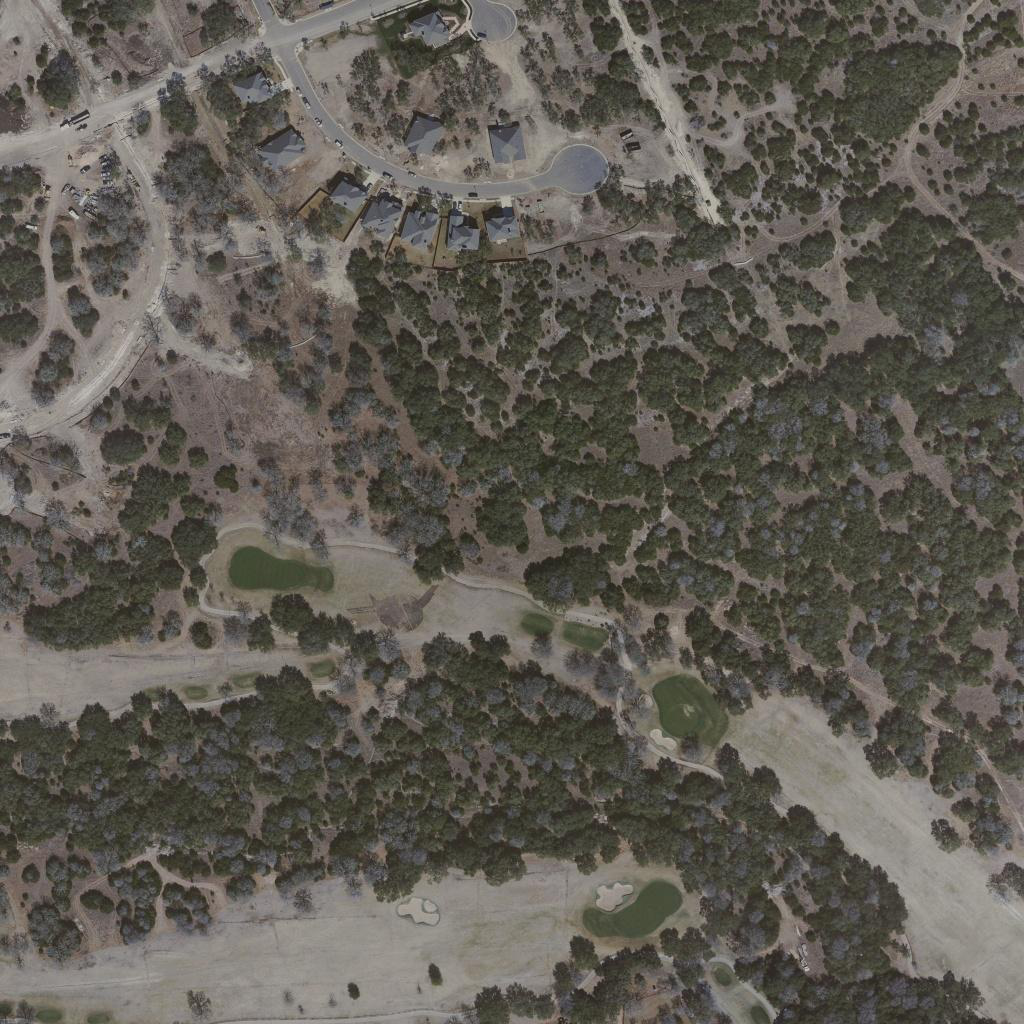}}
		{\includegraphics[width=.11\linewidth]{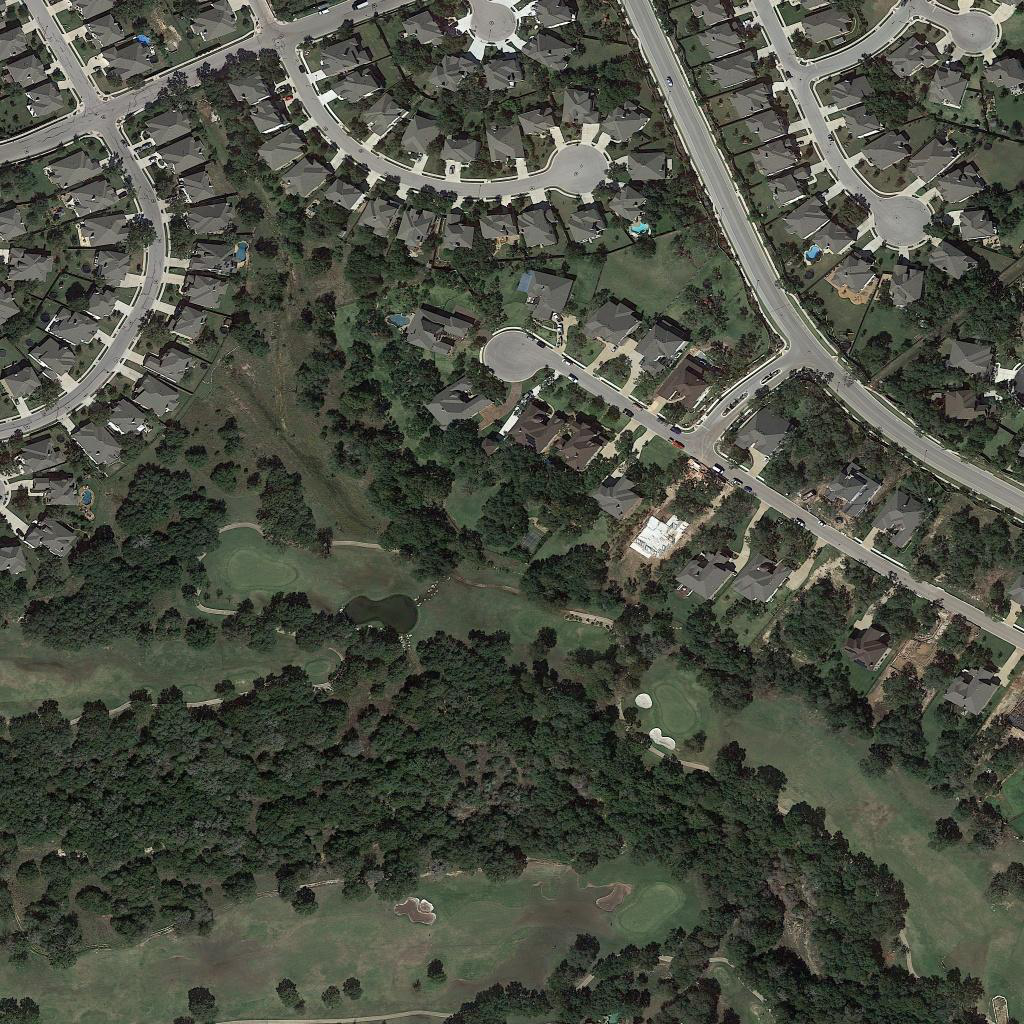}}
		{\includegraphics[width=.11\linewidth]{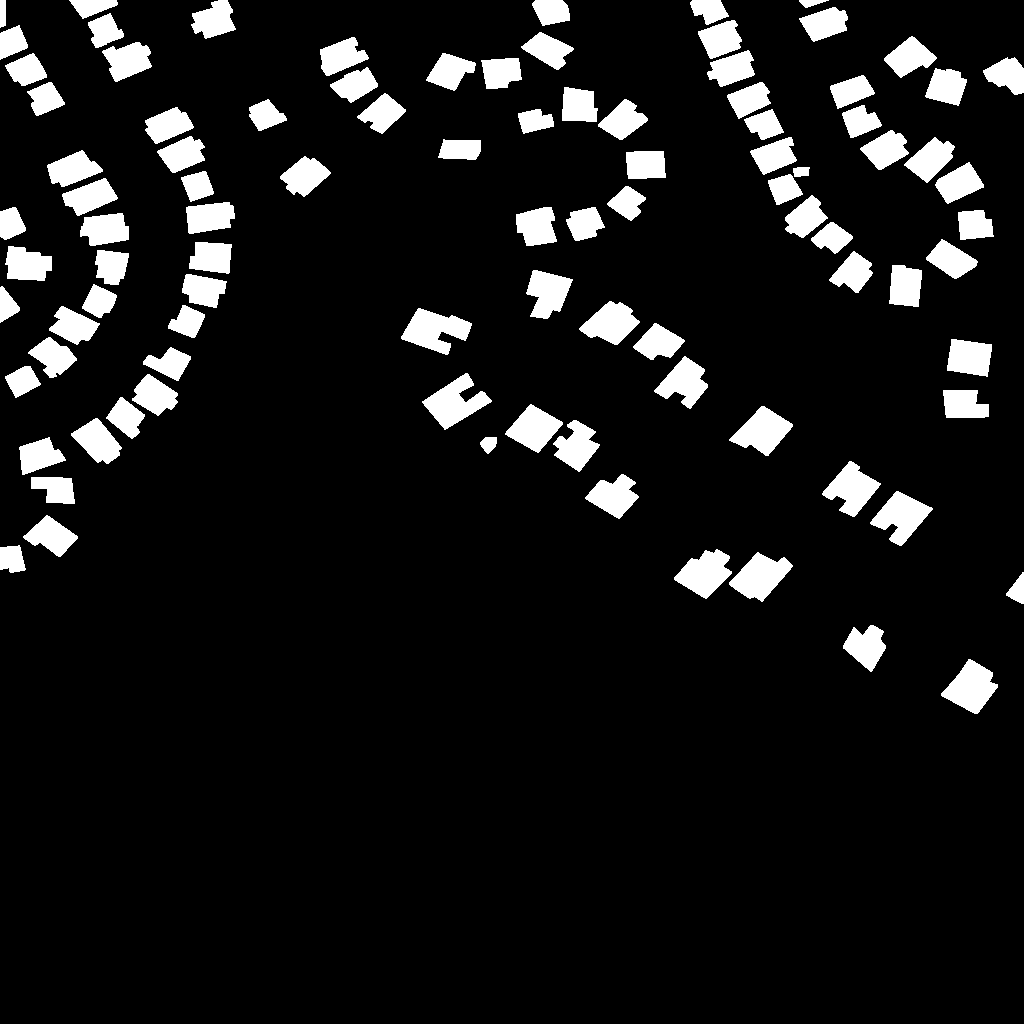}}
		{\includegraphics[width=.11\linewidth]{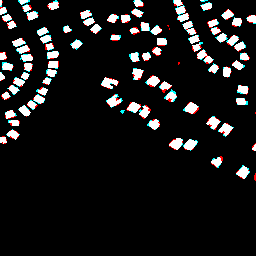}}
		{\includegraphics[width=.11\linewidth]{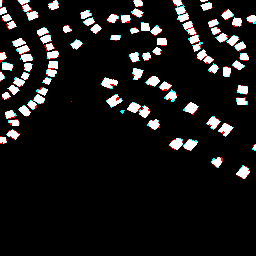}}
		{\includegraphics[width=.11\linewidth]{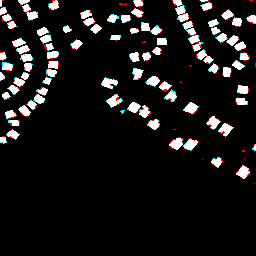}}
		{\includegraphics[width=.11\linewidth]{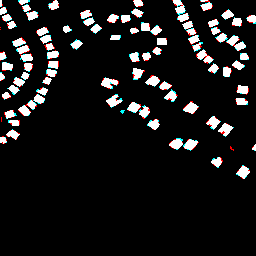}}
		{\includegraphics[width=.11\linewidth]{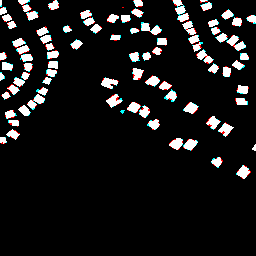}}
	\end{minipage}

	\vspace{0.1cm}

	\begin{minipage}{\linewidth}
		\centering
		{\includegraphics[width=.11\linewidth]{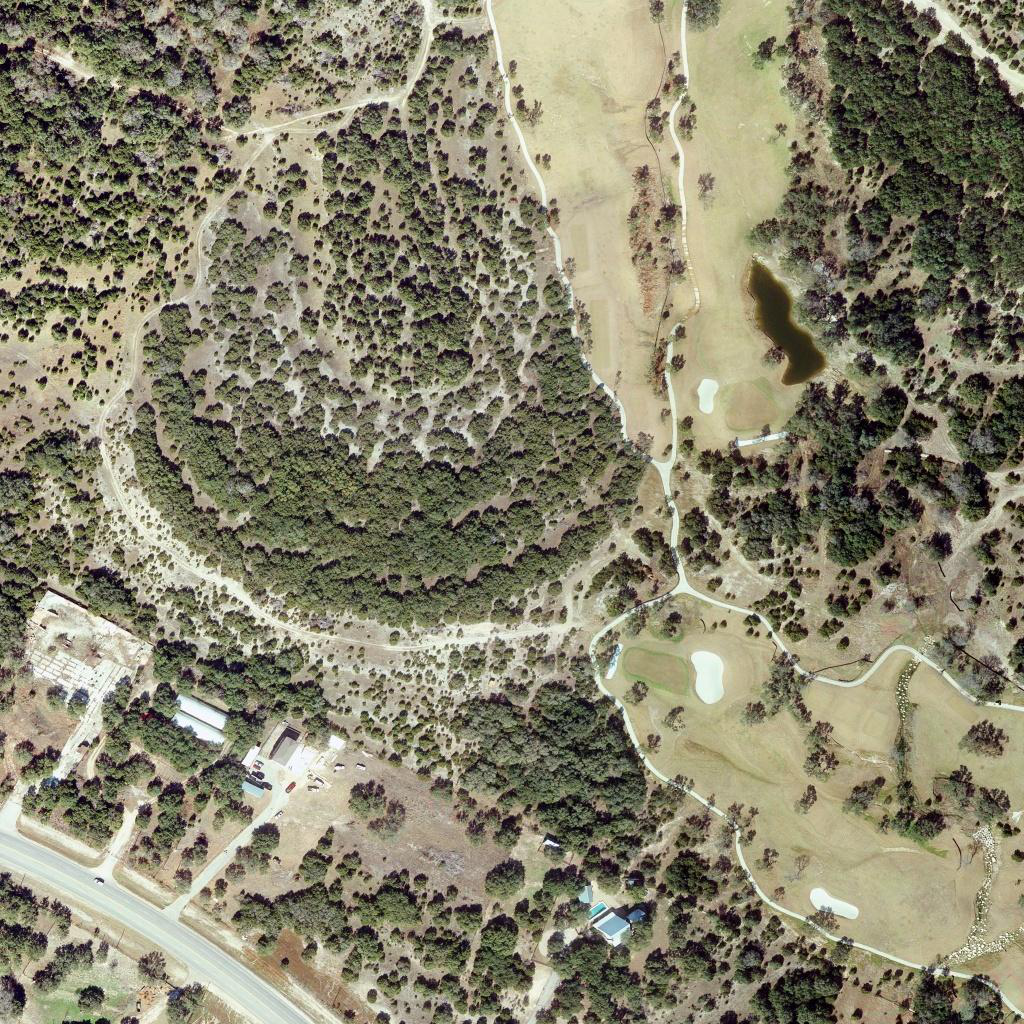}}
		{\includegraphics[width=.11\linewidth]{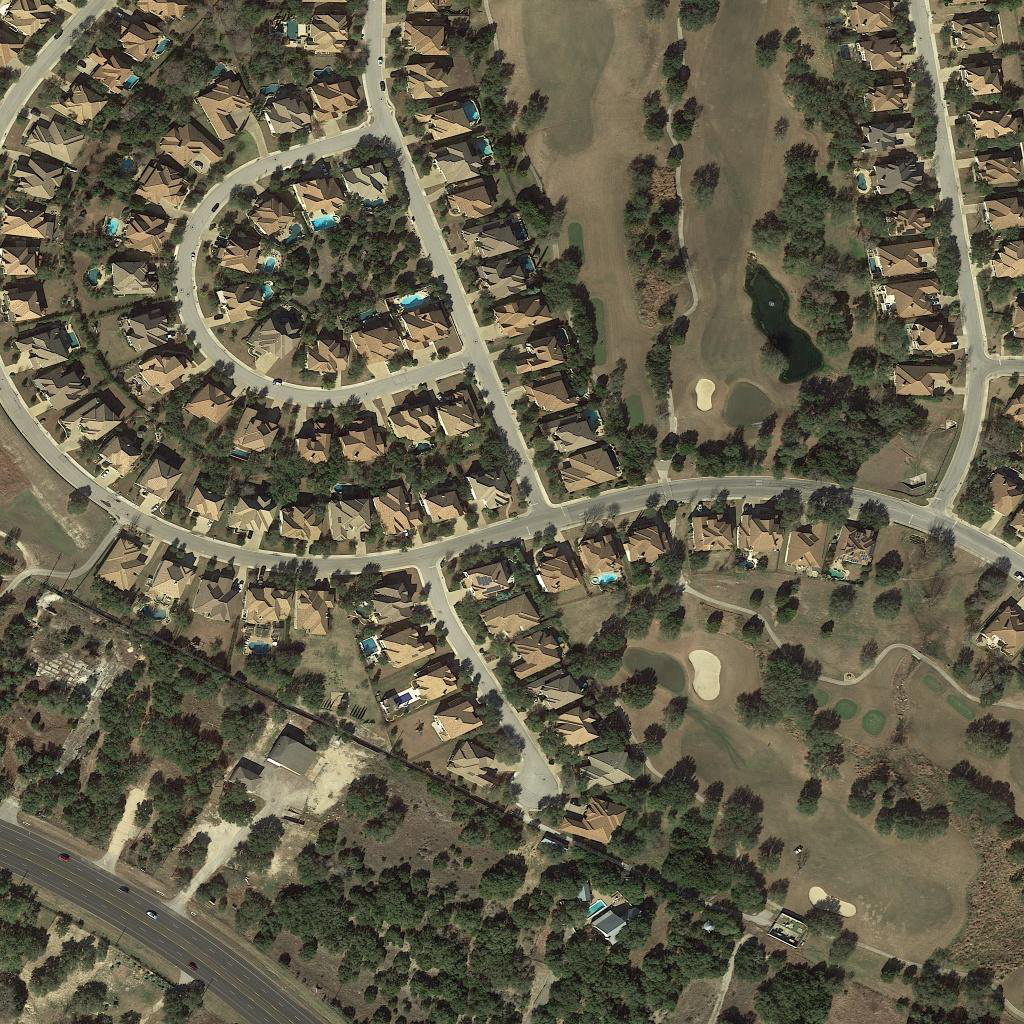}}
		{\includegraphics[width=.11\linewidth]{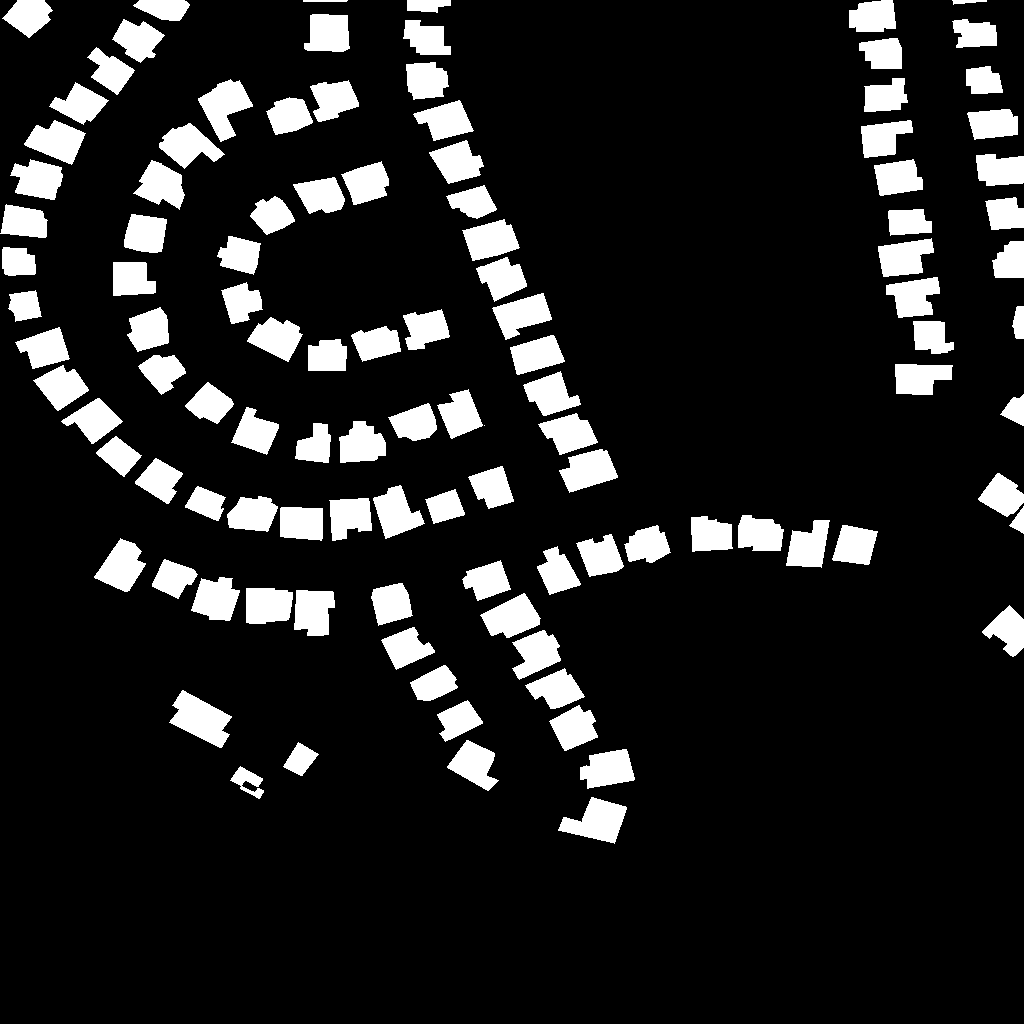}}
		{\includegraphics[width=.11\linewidth]{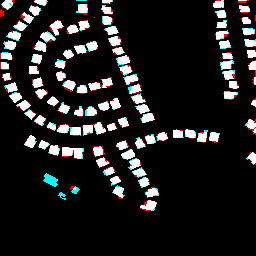}}
		{\includegraphics[width=.11\linewidth]{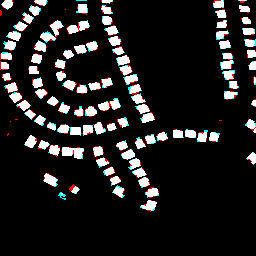}}
		{\includegraphics[width=.11\linewidth]{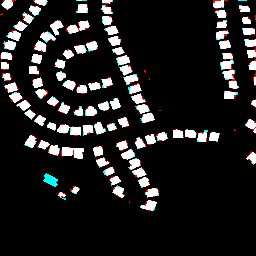}}
		{\includegraphics[width=.11\linewidth]{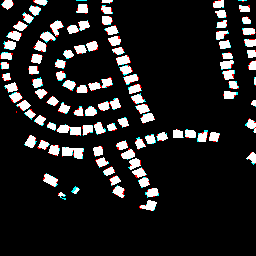}}
		{\includegraphics[width=.11\linewidth]{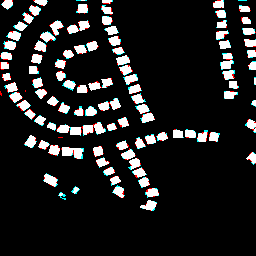}}
	\end{minipage}

	\vspace{0.1cm}

	\begin{minipage}{\linewidth}
		\centering
		{\includegraphics[width=.11\linewidth]{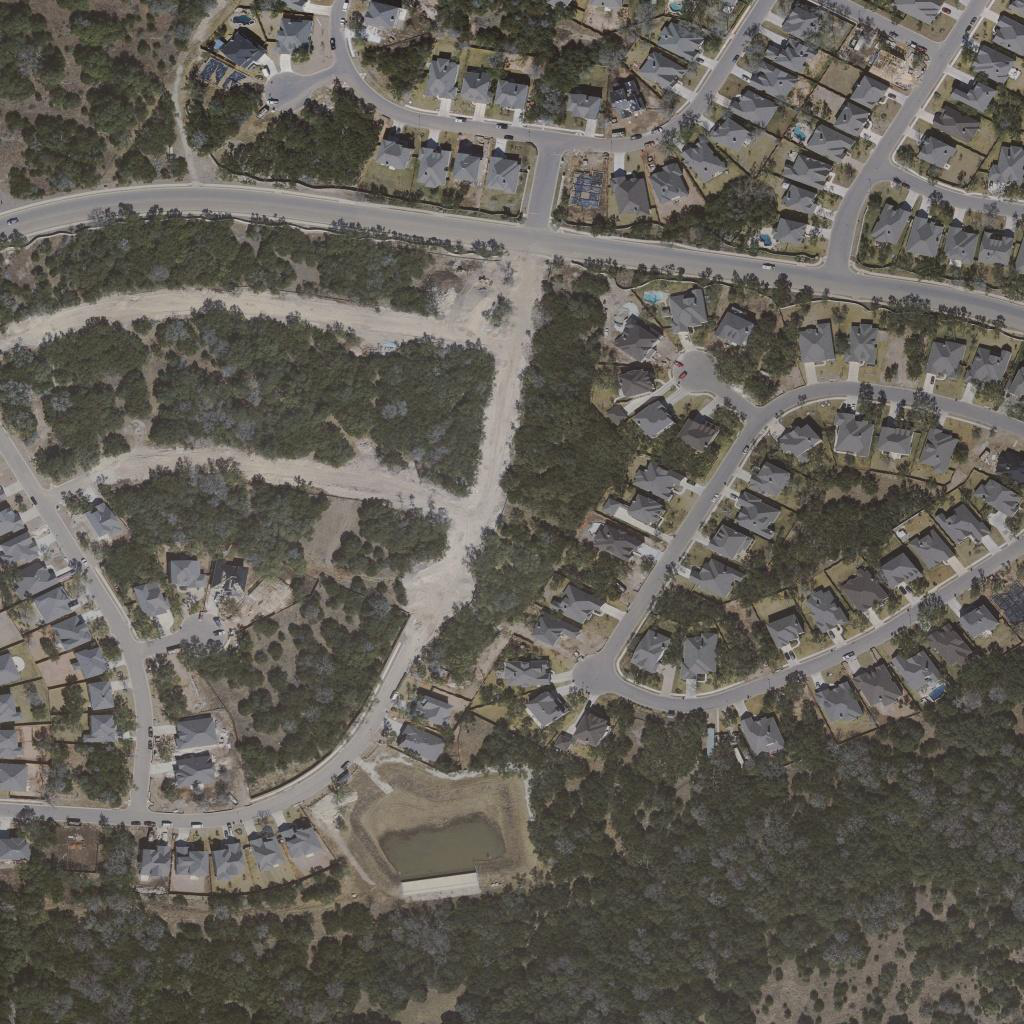}}
		{\includegraphics[width=.11\linewidth]{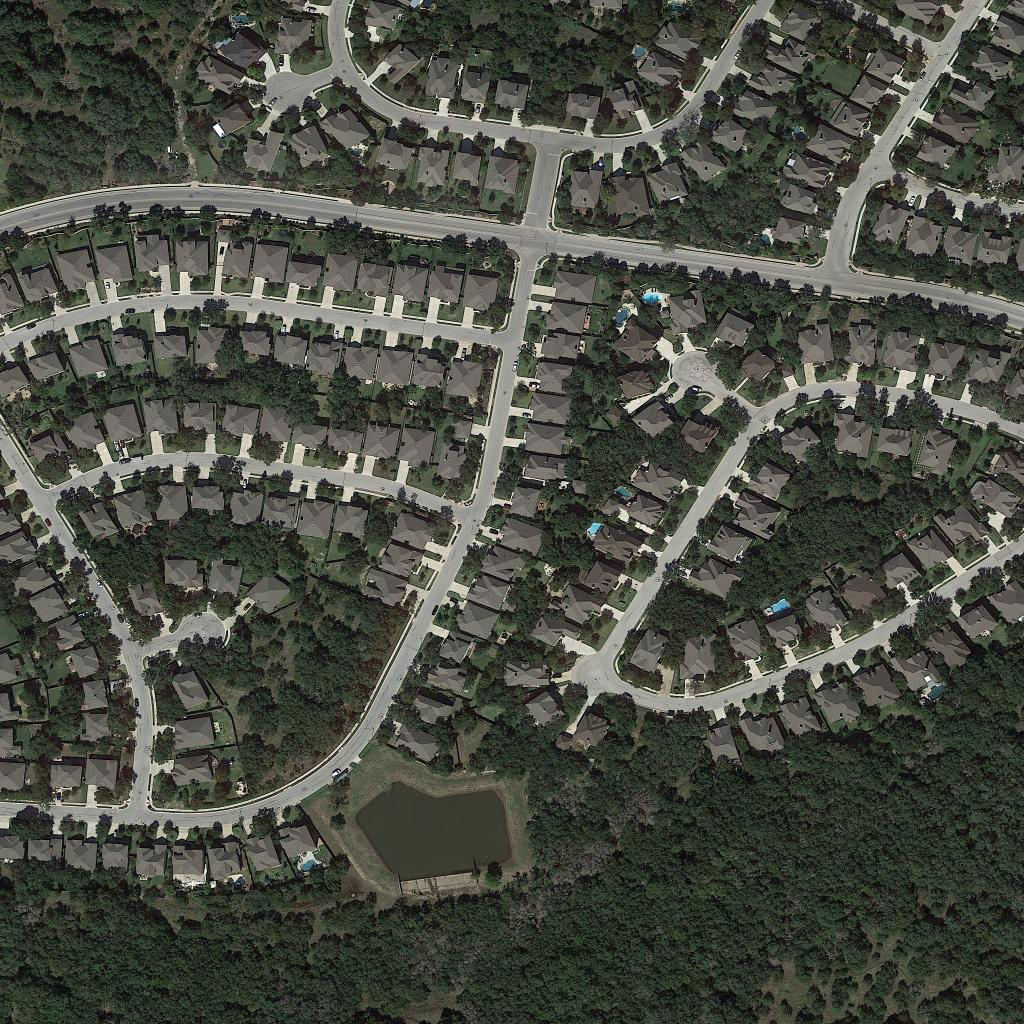}}
		{\includegraphics[width=.11\linewidth]{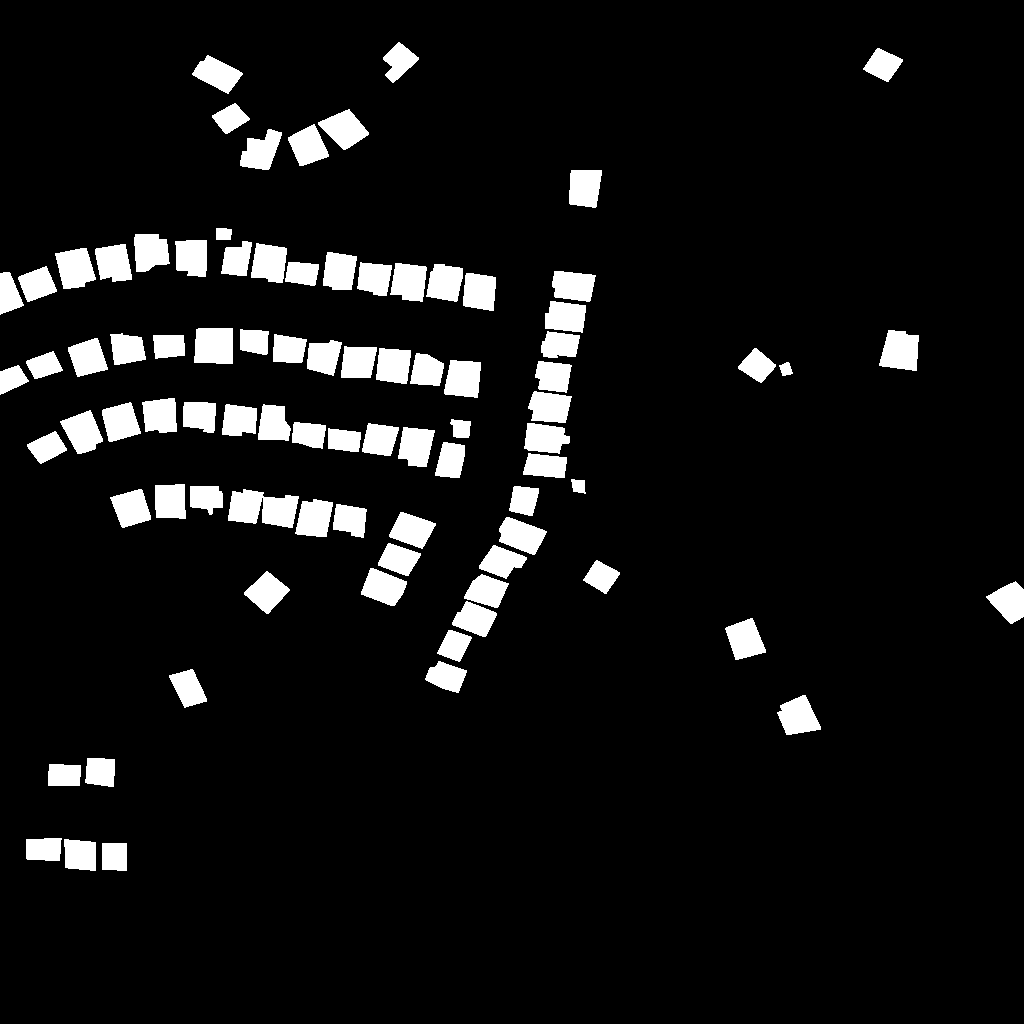}}
		{\includegraphics[width=.11\linewidth]{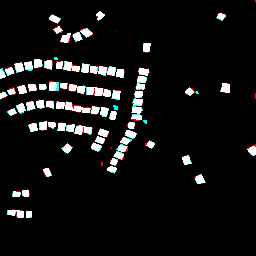}}
		{\includegraphics[width=.11\linewidth]{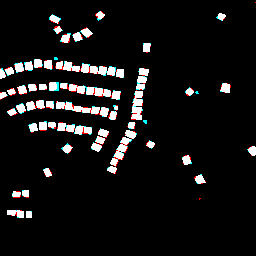}}
		{\includegraphics[width=.11\linewidth]{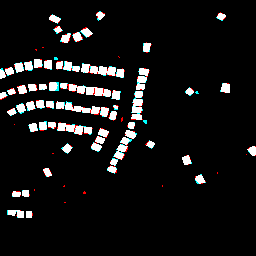}}
		{\includegraphics[width=.11\linewidth]{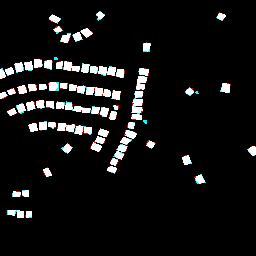}}
		{\includegraphics[width=.11\linewidth]{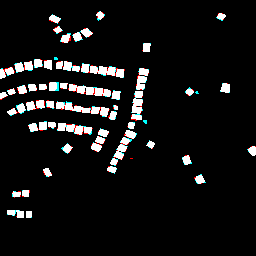}}
	\end{minipage}
	
	\vspace{0.1cm}
	
	\begin{minipage}{\linewidth}
		\setlength\tabcolsep{1.8pt}
		\begin{tabular}{p{.11\linewidth}<\centering p{.11\linewidth}<\centering p{.11\linewidth}<\centering p{.11\linewidth}<\centering p{.11\linewidth}<\centering p{.11\linewidth}<\centering p{.11\linewidth}<\centering p{.11\linewidth}<\centering p{.11\linewidth}<\centering}
			\qquad	\qquad	(a) & \qquad \qquad(b)  &  \;  \quad \qquad(c) &  \quad \qquad(d)&   \;  \qquad(e)  &  \quad \qquad(f) &   \quad \qquad(g) &  \quad \qquad(h) 
		\end{tabular}
	\end{minipage}
	\caption{Some samples of ablation experiments, where white indicates true positive, red indicates false positive, black indicates true negative, and blue indicates false negative. (a) T1 image. (a) T2 image. (c) Label. (d) DAGAN-R. (e) DAGAN-A. (f) DAGAN-M. (g) DAGAN-MC. (h) DAGAN.}
	\label{fig7}
\end{figure*}

\begin{table*}[htb]
	\centering
	\caption{Comparison experiments between the DAGAN and other state-of-the-art methods on the LEVIR test set. 
		\label{tab2}}
	\begin{tabular}{@{}ccccccc@{}}
		\toprule
		\multirow{2}{*}{Method} &\multirow{2}{*}{$Kappa$} & \multirow{2}{*}{$IoU$} & \multirow{2}{*}{$F1$}  & \multirow{2}{*}{$Recall$} & \multirow{2}{*}{$Prec$}   & \multirow{2}{*}{$Params$} \\
		&  &                        &                        &                 &    &  \\
    \midrule  FC-EF \cite{72}  &  80.63\%  & 68.84\%    & 81.54\%   & 83.55\%   & 79.63\%  & 1.35M    \\
        FC-Siam-cat \cite{72} &  84.62\%  & 74.43\%    & 85.34\%   & 86.93\%   & 83.81\%   & 1.55M   \\
        FC-Siam-diff \cite{72} &    86.49\%   &    77.18\%  &  87.12\%  & 88.14\%   &   86.12\%   & 1.35M    \\
        STANet \cite{73} &    88.48\%  & 80.21\%   & 89.02\%    & 90.01\%    & 88.05\%   & 12.28M  \\
        L-Unet \cite{74} &  89.02\%	   &  81.03\%         &    89.52\%       &    89.76\%       &       89.29\%    &       8.45M      \\
        SNUNet \cite{75}  &  89.55\%  &    81.86\%       &     90.03\%     &    89.17\%       &      90.90\%     &      27.07M       \\
       DSIFN \cite{76}  &    90.63\%  & 83.60\%   & 91.07\%   & 91.62\%   & 90.52\%   & 50.71M   \\
       BITNet \cite{77} &  90.59\%  & 83.53\%   & 91.02\%   & 91.31\%   & 90.74\%   & 3.01M   \\
       \textbf{Our}   &  \textbf{91.12\%}  &  \textbf{85.01\%}  &  \textbf{91.48\%}  &  \textbf{91.26\%}  &  \textbf{91.69\%}  &   \textbf{36.76M}  \\
  \bottomrule
	\end{tabular}
\end{table*}

\begin{figure*}[htbp]
	\centering
 
	\begin{minipage}{\linewidth}
		\centering
		{\includegraphics[width=.077\linewidth]{figs/fig7_1_1.png}}
		{\includegraphics[width=.077\linewidth]{figs/fig7_2_1.png}}
		{\includegraphics[width=.077\linewidth]{figs/fig7_3_1.png}}
		{\includegraphics[width=.077\linewidth]{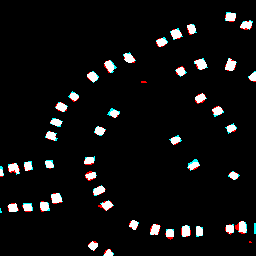}}
		{\includegraphics[width=.077\linewidth]{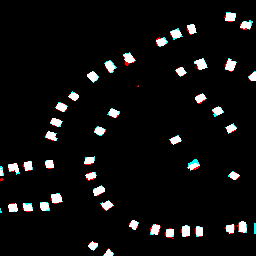}}
		{\includegraphics[width=.077\linewidth]{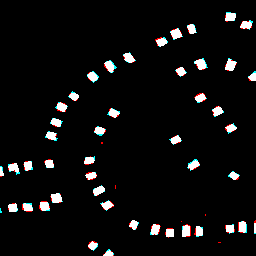}}
		{\includegraphics[width=.077\linewidth]{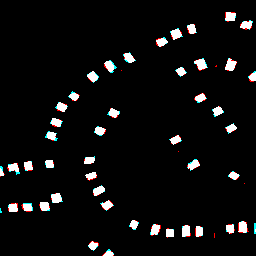}}
  		{\includegraphics[width=.077\linewidth]{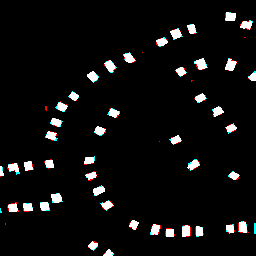}}
		{\includegraphics[width=.077\linewidth]{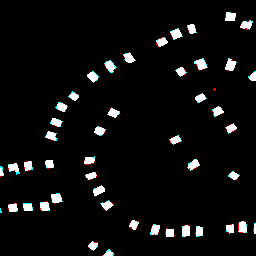}}
		{\includegraphics[width=.077\linewidth]{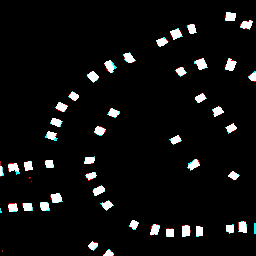}}
		{\includegraphics[width=.077\linewidth]{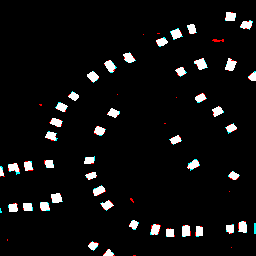}}
		{\includegraphics[width=.077\linewidth]{figs/fig7_8_1.png}}
	\end{minipage}
	
	\vspace{0.1cm}

		\begin{minipage}{\linewidth}
		\centering
		{\includegraphics[width=.077\linewidth]{figs/fig7_1_2.png}}
		{\includegraphics[width=.077\linewidth]{figs/fig7_2_2.png}}
		{\includegraphics[width=.077\linewidth]{figs/fig7_3_2.png}}
		{\includegraphics[width=.077\linewidth]{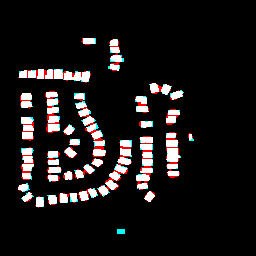}}
		{\includegraphics[width=.077\linewidth]{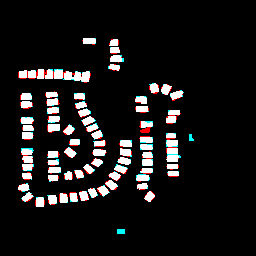}}
		{\includegraphics[width=.077\linewidth]{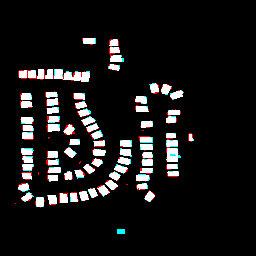}}
		{\includegraphics[width=.077\linewidth]{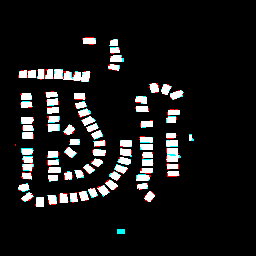}}
  		{\includegraphics[width=.077\linewidth]{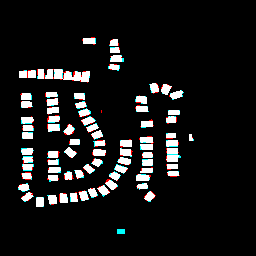}}
		{\includegraphics[width=.077\linewidth]{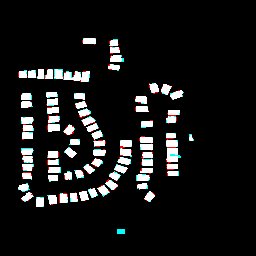}}
		{\includegraphics[width=.077\linewidth]{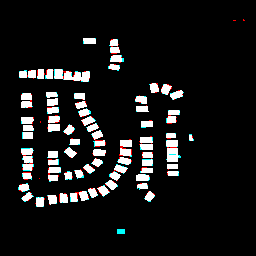}}
		{\includegraphics[width=.077\linewidth]{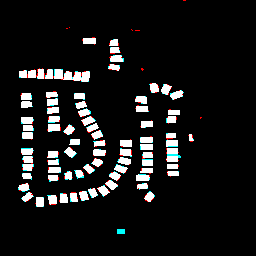}}
		{\includegraphics[width=.077\linewidth]{figs/fig7_8_2.png}}
	\end{minipage}
	
	\vspace{0.1cm}

		\begin{minipage}{\linewidth}
		\centering
		{\includegraphics[width=.077\linewidth]{figs/fig7_1_3.png}}
		{\includegraphics[width=.077\linewidth]{figs/fig7_2_3.png}}
		{\includegraphics[width=.077\linewidth]{figs/fig7_3_3.png}}
		{\includegraphics[width=.077\linewidth]{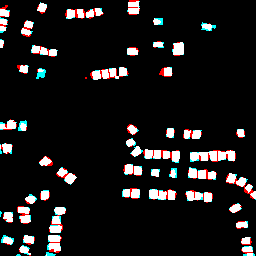}}
		{\includegraphics[width=.077\linewidth]{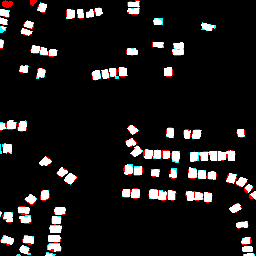}}
		{\includegraphics[width=.077\linewidth]{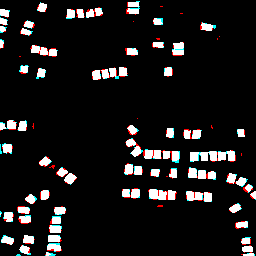}}
		{\includegraphics[width=.077\linewidth]{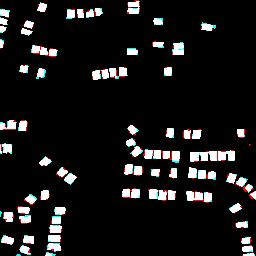}}
  		{\includegraphics[width=.077\linewidth]{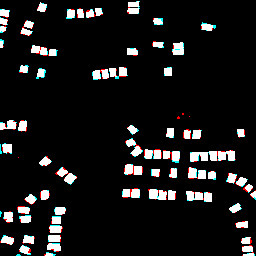}}
		{\includegraphics[width=.077\linewidth]{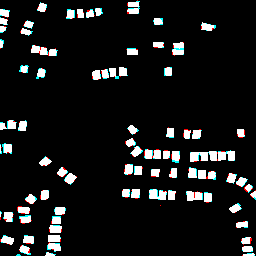}}
		{\includegraphics[width=.077\linewidth]{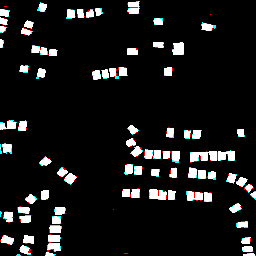}}
		{\includegraphics[width=.077\linewidth]{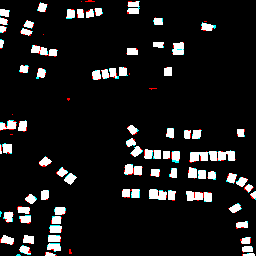}}
		{\includegraphics[width=.077\linewidth]{figs/fig7_8_1.png}}
	\end{minipage}

	\vspace{0.1cm}

		\begin{minipage}{\linewidth}
		\centering
		{\includegraphics[width=.077\linewidth]{figs/fig7_1_4.png}}
		{\includegraphics[width=.077\linewidth]{figs/fig7_2_4.png}}
		{\includegraphics[width=.077\linewidth]{figs/fig7_3_4.png}}
		{\includegraphics[width=.077\linewidth]{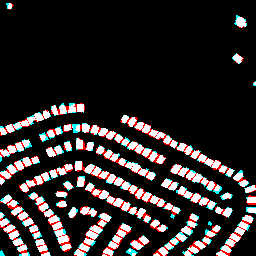}}
		{\includegraphics[width=.077\linewidth]{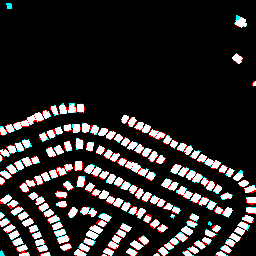}}
		{\includegraphics[width=.077\linewidth]{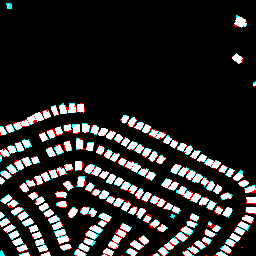}}
		{\includegraphics[width=.077\linewidth]{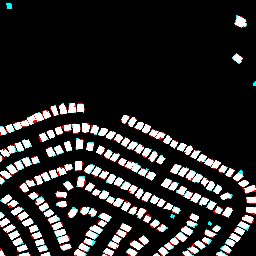}}
  		{\includegraphics[width=.077\linewidth]{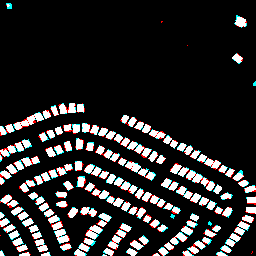}}
		{\includegraphics[width=.077\linewidth]{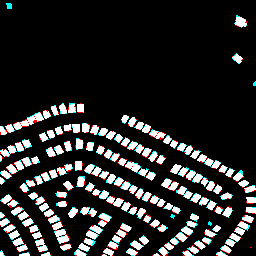}}
		{\includegraphics[width=.077\linewidth]{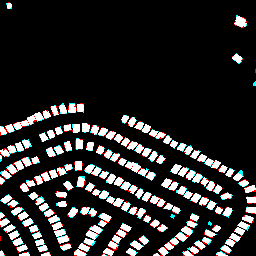}}
		{\includegraphics[width=.077\linewidth]{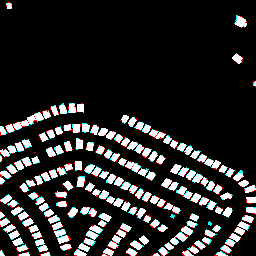}}
		{\includegraphics[width=.077\linewidth]{figs/fig7_8_4.png}}
	\end{minipage}

 	\vspace{0.1cm}

		\begin{minipage}{\linewidth}
		\centering
		{\includegraphics[width=.077\linewidth]{figs/fig7_1_5.png}}
		{\includegraphics[width=.077\linewidth]{figs/fig7_2_5.png}}
		{\includegraphics[width=.077\linewidth]{figs/fig7_3_5.png}}
		{\includegraphics[width=.077\linewidth]{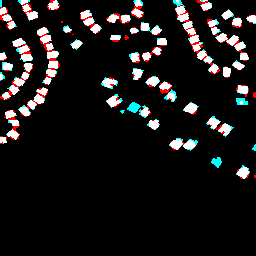}}
		{\includegraphics[width=.077\linewidth]{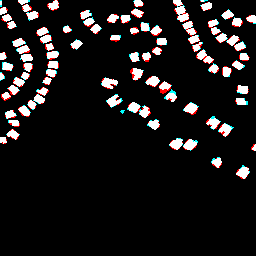}}
		{\includegraphics[width=.077\linewidth]{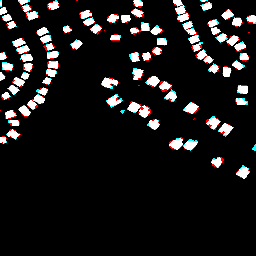}}
		{\includegraphics[width=.077\linewidth]{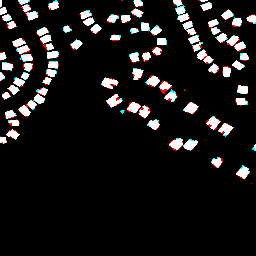}}
  		{\includegraphics[width=.077\linewidth]{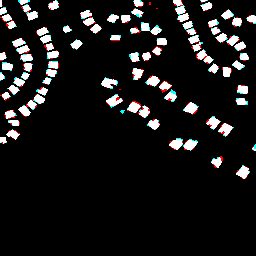}}
		{\includegraphics[width=.077\linewidth]{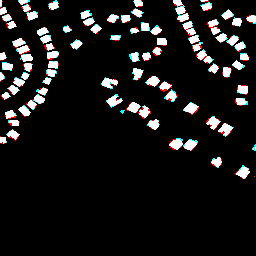}}
		{\includegraphics[width=.077\linewidth]{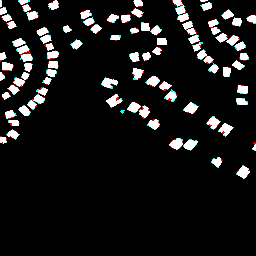}}
		{\includegraphics[width=.077\linewidth]{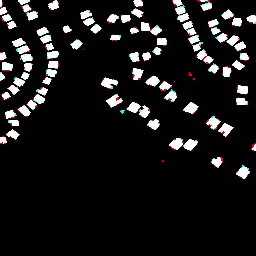}}
		{\includegraphics[width=.077\linewidth]{figs/fig7_8_5.png}}
	\end{minipage}

 	\vspace{0.1cm}

		\begin{minipage}{\linewidth}
		\centering
		{\includegraphics[width=.077\linewidth]{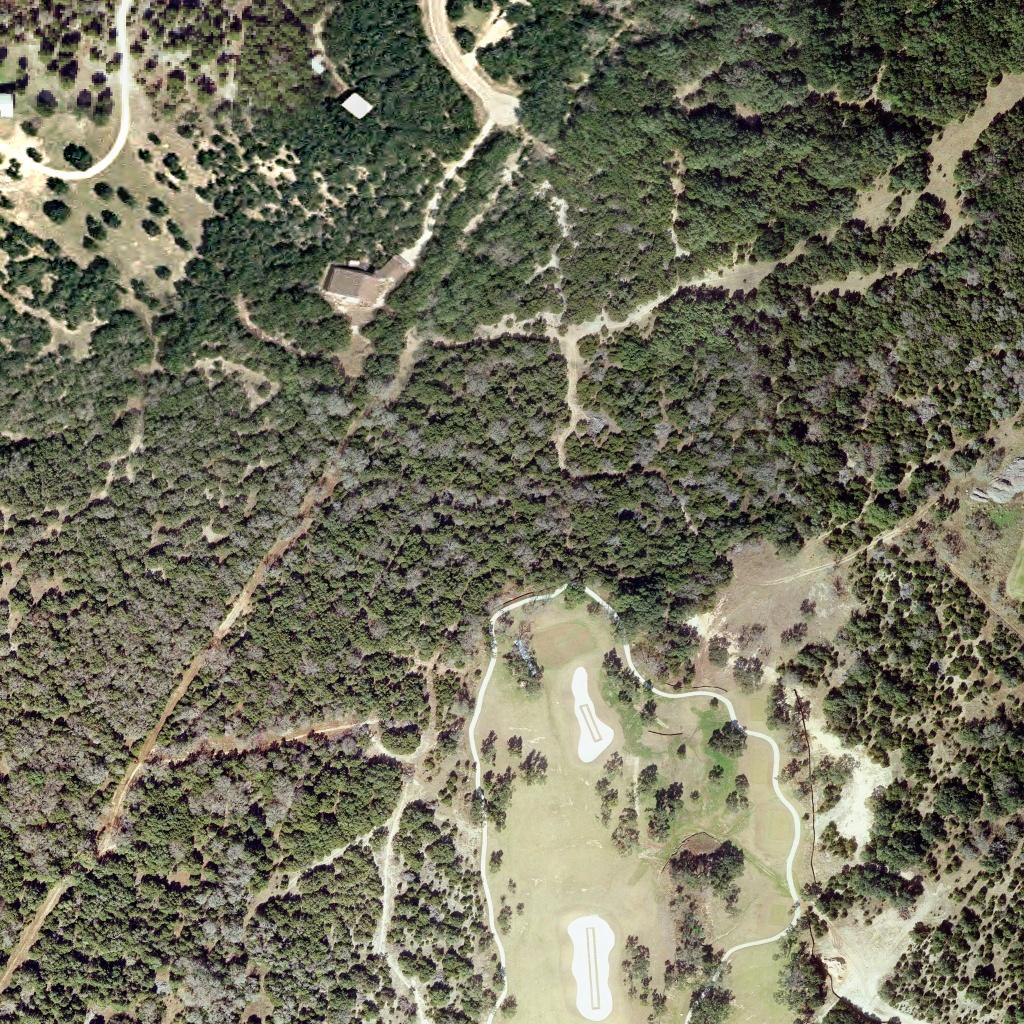}}
		{\includegraphics[width=.077\linewidth]{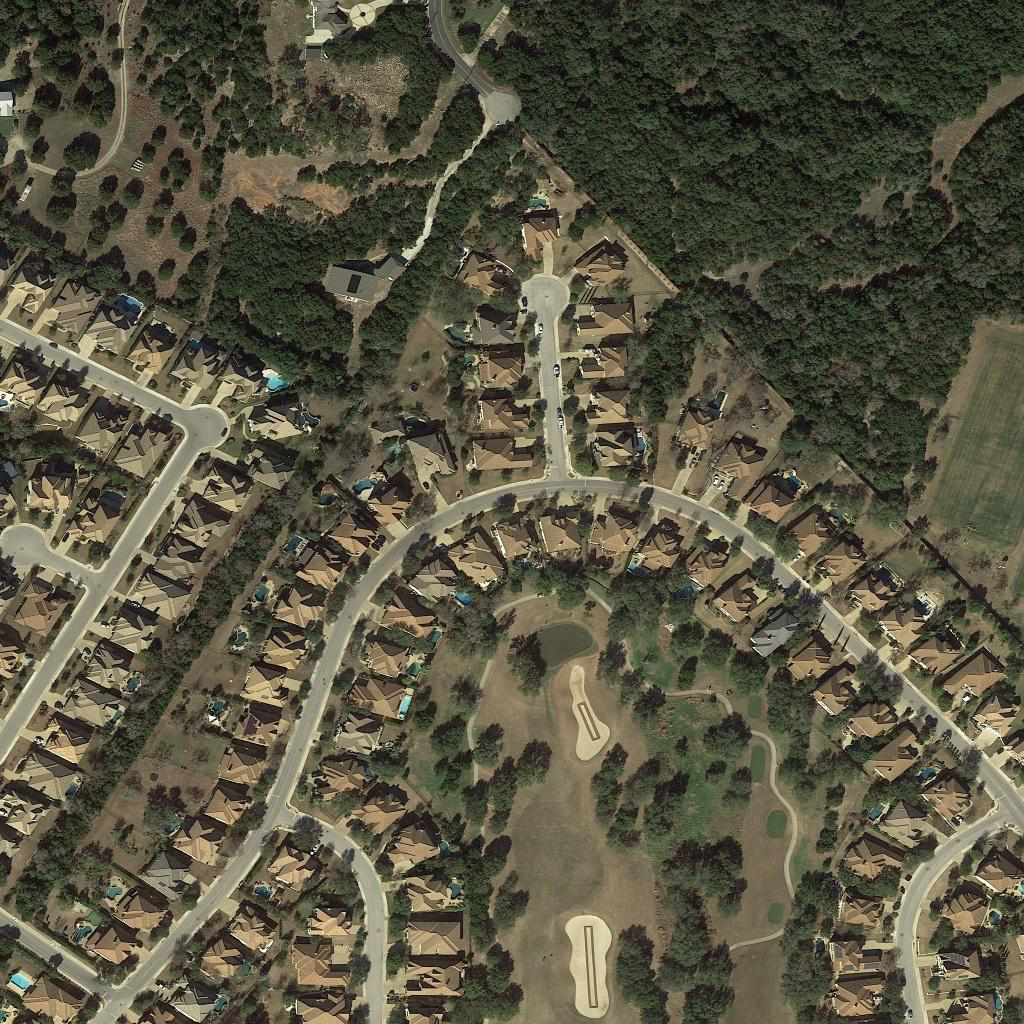}}
		{\includegraphics[width=.077\linewidth]{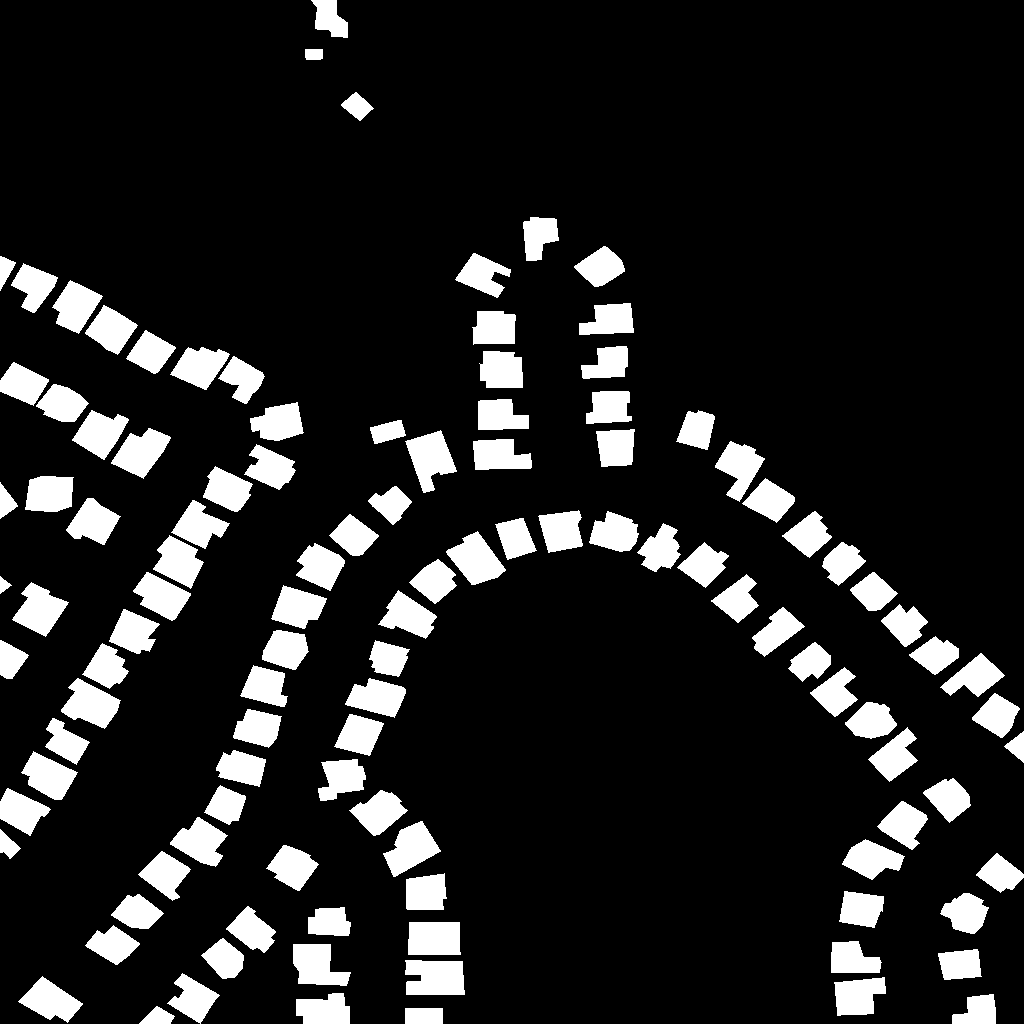}}
		{\includegraphics[width=.077\linewidth]{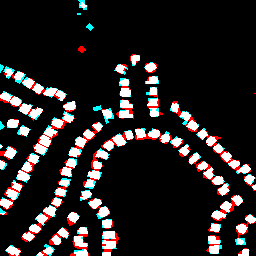}}
		{\includegraphics[width=.077\linewidth]{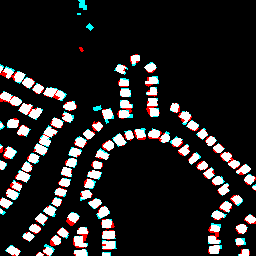}}
		{\includegraphics[width=.077\linewidth]{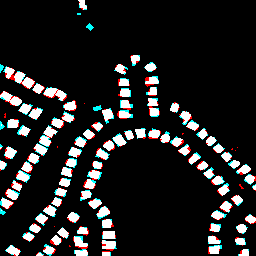}}
		{\includegraphics[width=.077\linewidth]{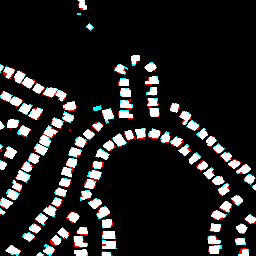}}
  		{\includegraphics[width=.077\linewidth]{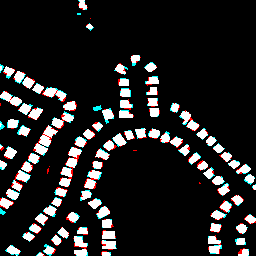}}
		{\includegraphics[width=.077\linewidth]{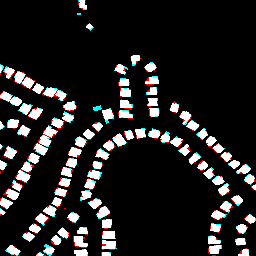}}
		{\includegraphics[width=.077\linewidth]{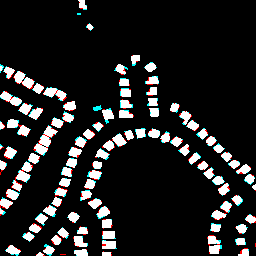}}
		{\includegraphics[width=.077\linewidth]{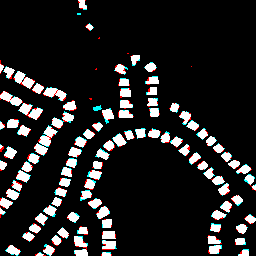}}
		{\includegraphics[width=.077\linewidth]{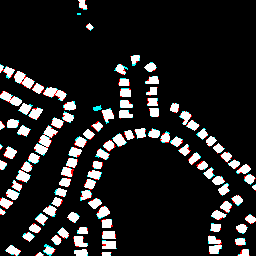}}
	\end{minipage}

 	\vspace{0.1cm}

		\begin{minipage}{\linewidth}
		\centering
		{\includegraphics[width=.077\linewidth]{figs/fig7_1_7.png}}
		{\includegraphics[width=.077\linewidth]{figs/fig7_2_7.png}}
		{\includegraphics[width=.077\linewidth]{figs/fig7_3_7.png}}
		{\includegraphics[width=.077\linewidth]{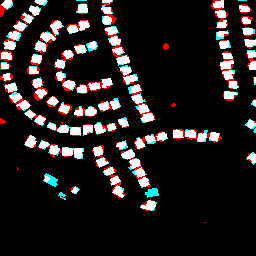}}
		{\includegraphics[width=.077\linewidth]{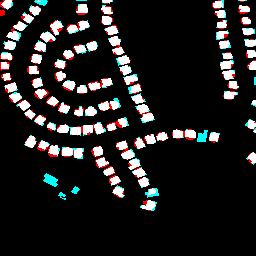}}
		{\includegraphics[width=.077\linewidth]{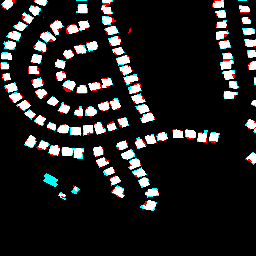}}
		{\includegraphics[width=.077\linewidth]{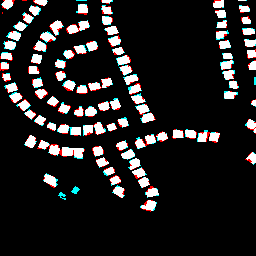}}
  		{\includegraphics[width=.077\linewidth]{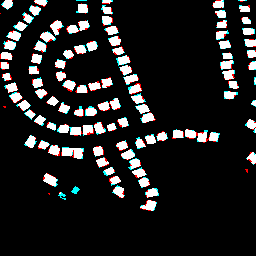}}
		{\includegraphics[width=.077\linewidth]{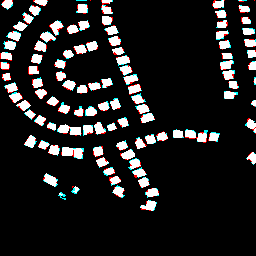}}
		{\includegraphics[width=.077\linewidth]{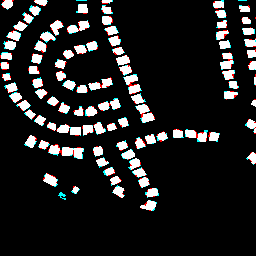}}
		{\includegraphics[width=.077\linewidth]{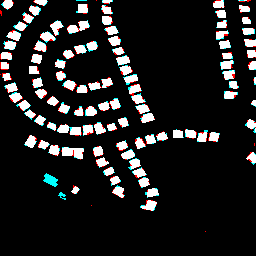}}
		{\includegraphics[width=.077\linewidth]{figs/fig7_8_7.png}}
	\end{minipage}

 	\vspace{0.1cm}

		\begin{minipage}{\linewidth}
		\centering
		{\includegraphics[width=.077\linewidth]{figs/fig7_1_8.png}}
		{\includegraphics[width=.077\linewidth]{figs/fig7_2_8.png}}
		{\includegraphics[width=.077\linewidth]{figs/fig7_3_8.png}}
		{\includegraphics[width=.077\linewidth]{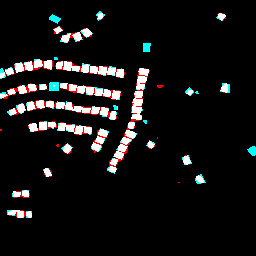}}
		{\includegraphics[width=.077\linewidth]{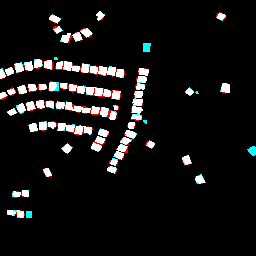}}
		{\includegraphics[width=.077\linewidth]{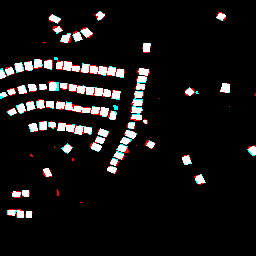}}
		{\includegraphics[width=.077\linewidth]{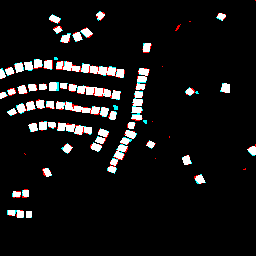}}
  		{\includegraphics[width=.077\linewidth]{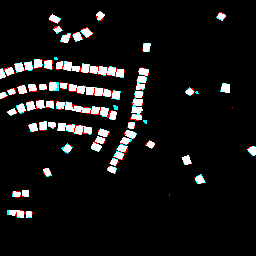}}
		{\includegraphics[width=.077\linewidth]{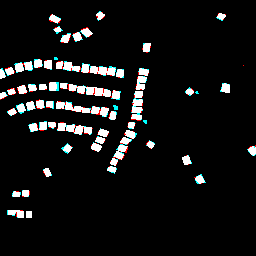}}
		{\includegraphics[width=.077\linewidth]{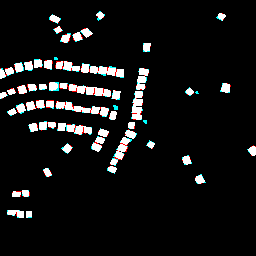}}
		{\includegraphics[width=.077\linewidth]{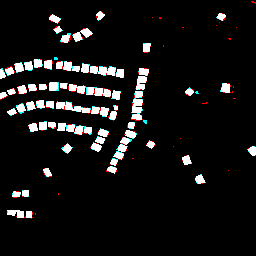}}
		{\includegraphics[width=.077\linewidth]{figs/fig7_8_8.png}}
	\end{minipage}

	\vspace{0.1cm}
	
	\begin{minipage}{\linewidth}
		\setlength\tabcolsep{1pt}
		\begin{tabular}{p{.077\linewidth}<\centering p{.077\linewidth}<\centering p{.077\linewidth}<\centering p{.077\linewidth}<\centering p{.077\linewidth}<\centering p{.077\linewidth}<\centering p{.077\linewidth}<\centering p{.077\linewidth}<\centering p{.077\linewidth}<\centering p{.077\linewidth}<\centering p{.077\linewidth}<\centering p{.077\linewidth}<\centering p{.077\linewidth}<\centering}
	\;	(a) & \; (b)  &  \;  (c) & \; (d)&   \;  (e)  & \; \quad (f) & \; \quad (g) & \; \qquad  (h) & \; \qquad  (i)  & \; \qquad (j) & \;  \qquad (k) & \; \qquad  (l) 
		\end{tabular}
	\end{minipage}
	\caption{Some samples of ablation experiments, where white indicates true positive, red indicates false positive, black indicates true negative, and blue indicates false negative.  (a) T1 image. (b) T2 image. (c) Label. (d) FC-EF \cite{72}. (e) FC-Siam-cat \cite{72}. (f) FC-Siam-diff \cite{72}. (g) STANet \cite{73}. (h) L-Unet \cite{74}. (i) SNUNet \cite{75}. (j) DSIFN \cite{76} (k) BITNet \cite{77}. (l) DAGAN.}
	\label{fig07}
\end{figure*}
We disintegrate the DAGAN framework to illustrate the effect of innovation parts. Each model is evaluated with the LEVIR dataset. We compare $kappa$, $IoU$, $f1\ score$, $recall$, $precision$ of each model, as shown in Table \ref{tab1}.

DAGAN-R. We use the pretraining ResNet50 models to extract initial features from images at the T1 and T2 moments, and then the framework fuses these features through the add operation. The $kappa$ and $IoU$ for the LEVIR test set are 88.17\% and 79.71\%.

DAGAN-A. We introduce aggregate connections to combine features from adjacent modules extracted by the pretraining model through the add operation. Compared with DAGAN-R, the $kappa$ and $IoU$ for the LEVIR test set are increased by 1.44\% and 2.26\%, and the $f1\ score$ is raised by 1.38\%.

DAGAN-M. We fuse features extracted from images at the T1 and T2 moments through four multi-scale adaptive fusion modules (MAFM). Compared with DAGAN-M, the $kappa$ and $IoU$ for the LEVIR test set are increased by 0.74\% and 1.18\%.

DAGAN-MC. After applying four multi-scale adaptive fusion modules (MAFM) to fuse features, we also design context refinement modules (CRM) to capture multi-scale features. The $IoU$ and $f1 \ score $ for the LEVIR test set are 84.75\% and 91.33\%.

DAGAN. We apply the generative adversarial strategy to optimize the DAGAN-MC framework, which could enhance the spatial contiguity of predictions. The $IoU$ and $f1 \ score $ for the LEVIR test set are 85.01\% and 91.48\%.

Figure \ref{fig6} is the visualization results of ablation experiments. The first and second column images collate at the T1 moment and T2 moment. The third column images label building differences of the T1 moment and T2 moment. The fourth column is the results of the DAGAN-R. The fifth column images predicted by the DAGAN-A, and the sixth and seventh column images generated by the DAGAN-M and the DAGAN-MC, respectively. And the eighth column images predicted by the DAGAN. Compared with other framework, the continuity and completeness of the buildings is significantly improved, which shows that aggregate connection, MAFM and CFM modules could effectively improve the recognition ability of the framework.

\subsection{Comparison Experiments} 

The paper chooses some classic and advanced methods for change detection tasks as comparisons methods. As shown in Table \ref{tab2}, FC-EF \cite{72} obtains worst detection performance than them, which might result from the loss of numerous differences through simply adding channel features from paired images. The $IoU$ and $f1 score$ for the LEVIR test set are 68.17\% and 81.54\%. FC-Siam-cat \cite{72} concatenates difference features extracted from a dual-branch encoder and achieves better detection capability than FC-EF \cite{72}. The $IoU$ and $f1 score$ for the LEVIR test set are 74.43\% and 85.34\%. Compared with FC-Siam-cat \cite{72}, FC-Siam-diff \cite{72} concatenates the absolute value of difference features, and the $IoU$ and the $f1 score$ for the LEVIR test set increase to 77.18\% and 87.12\%. And the $IoU$ and $f1 score$ of STANet \cite{73} for the LEVIR test set are 80.21\% and 89.02\%. The $IoU$ and the $f1 score$ of L-Unet \cite{74} for the LEVIR test set increase to 81.03\% and 89.52\%. SNUNet \cite{75} refines difference features through ensemble channel attention modules, which alleviates the loss of localization information, and the $IoU$ and the $f1 score$ for the LEVIR test set improves to 81.86\% and 90.03\%. DSIFN \cite{76} extracts multi-level features from paired images through a two-stream architecture and enhances boundary completeness by adopting attention modules to fuse differences. The $IoU$ for the LEVIR test set is 83.60\%, and the $f1 score$ for the LEVIR test set is 91.07\%. BITNet \cite{77} introduces a transformer as an encoder to learn contextual information and refine features through a transformer decoder, which helps the framework to capture richer context features. The $IoU$ for the LEVIR test set is 83.53\%. Inspired by these fusion methods, the DAGAN introduces multi-scale adaptive fusion modules to aggregate multi-level features extracted by pre-trained models progressively and designs context refinement modules to capture differences from paired images, which has achieved significantly better than other advanced methods.

 Figure \ref{fig07} is the visualization of the comparison experiments. The first and second column images collate at the T1 moment and T2 moment. The third column results are the ground truth of the differences between the T1 and T2 moments. The fourth column is the results of FC-EF \cite{72}. The fifth and sixth column images are predicted by FC-Siam-cat and FC-Siam-diff \cite{72}. The seventh column results are generated by STANet \cite{73}. The eighth column results are predicted by L-Unet \cite{74}. The ninth column results are generated by SNUNet \cite{75}. The tenth and eleventh column results are predicted by DSIFN \cite{76}, and BITNet \cite{77}, respectively. The results of the twelfth column are predicted by the DAGAN, which has more complete boundaries.

\section{Conclusion}
\vspace{0.1cm}
By incorporating a generative-adversarial strategy, we propose a novel dual attentive generative-adversarial framework (DAGAN) for change detection tasks of VHR images. The DAGAN framework adopts a multi-level feature extractor to extract multi-level features from bi-temporal images by pre-trained models and introduces aggregate connections to fuse them. Moreover, the DAGAN framework utilizes a multi-scale adaptive fusion module with various receptive fields to fuse multi-scale features and a context refinement module to explore contextual dependencies, which could enhance the recognition ability of multi-scale objects.

\section{Acknowledgements} Many thanks to the Joint Fund of the Ministry of Education No.8091B022101 for their support and funding. In the course of our research, the support of this fund has played an important role in promoting and helping our work. This funding provides us with experimental materials, equipment, technical support, and the costs of conducting academic exchanges and publishing research results. Without these grants, our research would not be able to proceed as planned or achieve the expected results.


\bibliographystyle{splncs04}

\begin{thebibliography}{10}

\bibitem{78}
Beifang Bai, Wei Fu, Ting Lu, and Shutao Li.
\newblock Edge-guided recurrent convolutional neural network for multitemporal
  remote sensing image building change detection.
\newblock {\em IEEE Transactions on Geoscience and Remote Sensing}, 60:1--13,
  2021.

\bibitem{9540260}
Ricardo~Augusto Borsoi, Tales Imbiriba, José Carlos~Moreira Bermudez, and
  Cédric Richard.
\newblock Fast unmixing and change detection in multitemporal hyperspectral
  data.
\newblock {\em IEEE Transactions on Computational Imaging}, 7:975--988, 2021.

\bibitem{87}
Wonmin Byeon, Thomas~M. Breuel, Federico Raue, and Marcus Liwicki.
\newblock Scene labeling with {LSTM} recurrent neural networks.
\newblock In {\em Proceddings of the {IEEE} Conference on Computer Vision and
  Pattern Recognition}, pages 3547--3555, Boston, MA, USA, 2015. {IEEE}
  Computer Society.

\bibitem{51}
Hao Chen, Zipeng Qi, and Zhenwei Shi.
\newblock Remote sensing image change detection with transformers.
\newblock {\em IEEE Transactions on Geoscience and Remote Sensing}, 60:1--14,
  2021.

\bibitem{77}
Hao Chen, Zipeng Qi, and Zhenwei Shi.
\newblock Remote sensing image change detection with transformers.
\newblock {\em IEEE Transactions on Geoscience and Remote Sensing}, 60:1--14,
  2021.

\bibitem{73}
Hao Chen and Zhenwei Shi.
\newblock A spatial-temporal attention-based method and a new dataset for
  remote sensing image change detection.
\newblock {\em Remote Sensing}, 12(10):1662, 2020.

\bibitem{72}
Rodrigo~Caye Daudt, Bertr Le~Saux, and Alexandre Boulch.
\newblock Fully convolutional siamese networks for change detection.
\newblock In {\em 2018 25th IEEE International Conference on Image Processing
  (ICIP)}, pages 4063--4067. IEEE, 2018.

\bibitem{10004200}
Deepanshi, Rahasya Barkur, Devishi Suresh, Shyam Lal, C.~Sudhakar Reddy, and
  P.~G. Diwakar.
\newblock Rscdnet: A robust deep learning architecture for change detection
  from bi-temporal high resolution remote sensing images.
\newblock {\em IEEE Transactions on Emerging Topics in Computational
  Intelligence}, 7(2):537--551, 2023.

\bibitem{75}
Sheng Fang, Kaiyu Li, Jinyuan Shao, and Zhe Li.
\newblock Snunet-cd: A densely connected siamese network for change detection
  of vhr images.
\newblock {\em IEEE Geoscience and Remote Sensing Letters}, 19:1--5, 2021.

\bibitem{79}
Wenjie Geng, Zhiqiang Cao, Peiyu Guan, Guangli Ren, Junzhi Yu, and Fengshui
  Jing.
\newblock Adaptive long-neck network with atrous-residual structure for
  instance segmentation.
\newblock {\em IEEE Sensors Journal}, 23(7):7786--7797, 2023.

\bibitem{7}
Ian~J. Goodfellow, Jean Pouget{-}Abadie, Mehdi Mirza, Bing Xu, David
  Warde{-}Farley, Sherjil Ozair, Aaron~C. Courville, and Yoshua Bengio.
\newblock Generative adversarial nets.
\newblock In {\em Proceedings of the Advances in Neural Information Processing
  Systems 27: Annual Conference on Neural Information Processing Systems},
  pages 2672--2680, Montreal, Quebec, Canada, 2014.

\bibitem{8809352}
Neha Gupta, Samit Ari, and Narayan Panigrahi.
\newblock Change detection in landsat images using unsupervised learning and
  rbf-based clustering.
\newblock {\em IEEE Transactions on Emerging Topics in Computational
  Intelligence}, 5(2):284--297, 2021.

\bibitem{19}
Mohammad Hamghalam, Baiying Lei, and Tianfu Wang.
\newblock High tissue contrast {MRI} synthesis using multi-stage attention-gan
  for segmentation.
\newblock In {\em Proceedings of the Thirty-Fourth {AAAI} Conference on
  Artificial Intelligence}, pages 4067--4074, New York, NY, USA, 2020. {AAAI}
  Press.

\bibitem{89}
Renlong Hang, Ping Yang, Feng Zhou, and Qingshan Liu.
\newblock Multiscale progressive segmentation network for high-resolution
  remote sensing imagery.
\newblock {\em IEEE Transactions on Geoscience and Remote Sensing}, 60:1--12,
  2022.

\bibitem{70}
Feng Jiang, Yang Lu, Yu~Chen, Di~Cai, and Gongfa Li.
\newblock Image recognition of four rice leaf diseases based on deep learning
  and support vector machine.
\newblock {\em Computers and Electronics in Agriculture}, 179:105824, 2020.

\bibitem{91}
Xinghua Li, Meizhen He, Huifang Li, and Huanfeng Shen.
\newblock A combined loss-based multiscale fully convolutional network for
  high-resolution remote sensing image change detection.
\newblock {\em IEEE Geoscience and Remote Sensing Letters}, 19:1--5, 2021.

\bibitem{56}
Zhenglai Li, Chang Tang, Xinwang Liu, Wei Zhang, Jie Dou, Lizhe Wang, and
  Albert~Y Zomaya.
\newblock Lightweight remote sensing change detection with progressive feature
  aggregation and supervised attention.
\newblock {\em IEEE Transactions on Geoscience and Remote Sensing}, 61:1--12,
  2023.

\bibitem{54}
Ziming Li, Chenxi Yan, Ying Sun, and Qinchuan Xin.
\newblock A densely attentive refinement network for change detection based on
  very-high-resolution bitemporal remote sensing images.
\newblock {\em IEEE Transactions on Geoscience and Remote Sensing}, 60:1--18,
  2022.

\bibitem{9975266}
Manhui Lin, Guangyi Yang, and Hongyan Zhang.
\newblock Transition is a process: Pair-to-video change detection networks for
  very high resolution remote sensing images.
\newblock {\em IEEE Transactions on Image Processing}, 32:57--71, 2023.

\bibitem{8219717}
Zhunga Liu, Gang Li, Gregoire Mercier, You He, and Quan Pan.
\newblock Change detection in heterogenous remote sensing images via
  homogeneous pixel transformation.
\newblock {\em IEEE Transactions on Image Processing}, 27(4):1822--1834, 2018.

\bibitem{7422756}
Xiaoqiang Lu, Yuan Yuan, and Xiangtao Zheng.
\newblock Joint dictionary learning for multispectral change detection.
\newblock {\em IEEE Transactions on Cybernetics}, 47(4):884--897, 2017.

\bibitem{22}
Pauline Luc, Camille Couprie, Soumith Chintala, and Jakob Verbeek.
\newblock Semantic segmentation using adversarial networks.
\newblock In {\em NIPS Workshop on Adversarial Training}, 2016.

\bibitem{92}
Zhiyong Lv, Haitao Huang, Lipeng Gao, Jon~Atli Benediktsson, Minghua Zhao, and
  Cheng Shi.
\newblock Simple multiscale unet for change detection with heterogeneous remote
  sensing images.
\newblock {\em IEEE Geoscience and Remote Sensing Letters}, 19:1--5, 2022.

\bibitem{82}
Bifang Ma and Chih-Yung Chang.
\newblock Semantic segmentation of high-resolution remote sensing images using
  multiscale skip connection network.
\newblock {\em IEEE Sensors Journal}, 22(4):3745--3755, 2021.

\bibitem{71}
Lan~H Nguyen, Deepak~R Joshi, David~E Clay, and Geoffrey~M Henebry.
\newblock Characterizing land cover/land use from multiple years of landsat and
  modis time series: A novel approach using land surface phenology modeling and
  random forest classifier.
\newblock {\em Remote sensing of environment}, 238:111017, 2020.

\bibitem{65}
Yuzhen Niu, Jianbin Wu, Wenxi Liu, Wenzhong Guo, and Rynson~WH Lau.
\newblock Hdr-gan: Hdr image reconstruction from multi-exposed ldr images with
  large motions.
\newblock {\em IEEE Transactions on Image Processing}, 30:3885--3896, 2021.

\bibitem{74}
Maria Papadomanolaki, Maria Vakalopoulou, and Konstantinos Karantzalos.
\newblock A deep multitask learning framework coupling semantic segmentation
  and fully convolutional lstm networks for urban change detection.
\newblock {\em IEEE Transactions on Geoscience and Remote Sensing},
  59(9):7651--7668, 2021.

\bibitem{80}
Luyi Qiu, Dayu Yu, Chenxiao Zhang, and Xiaofeng Zhang.
\newblock A local--global framework for semantic segmentation of multisource
  remote sensing images.
\newblock {\em Remote Sensing}, 15(1):231, 2022.

\bibitem{81}
Luyi Qiu, Dayu Yu, Chenxiao Zhang, and Xiaofeng Zhang.
\newblock A semantics-geometry framework for road extraction from remote
  sensing images.
\newblock {\em IEEE Geoscience and Remote Sensing Letters}, 2023.

\bibitem{69}
Ronghua Shang, Kaize Xie, Michael~Aggrey Okoth, and Licheng Jiao.
\newblock Sar image change detection based on mean shift pre-classification and
  fuzzy c-means.
\newblock In {\em IGARSS 2019-2019 IEEE International Geoscience and Remote
  Sensing Symposium}, pages 2358--2361. IEEE, 2019.

\bibitem{52}
Qian Shi, Mengxi Liu, Shengchen Li, Xiaoping Liu, Fei Wang, and Liangpei Zhang.
\newblock A deeply supervised attention metric-based network and an open aerial
  image dataset for remote sensing change detection.
\newblock {\em IEEE transactions on geoscience and remote sensing}, 60:1--16,
  2021.

\bibitem{86}
Xiaolong Wang, Ross~B. Girshick, Abhinav Gupta, and Kaiming He.
\newblock Non-local neural networks.
\newblock In {\em Proceedings of the {IEEE} Conference on Computer Vision and
  Pattern Recognition}, pages 7794--7803, Salt Lake City, UT, USA, 2018.
  Computer Vision Foundation / {IEEE} Computer Society.

\bibitem{9477493}
Chen Wu, Hongruixuan Chen, Bo~Du, and Liangpei Zhang.
\newblock Unsupervised change detection in multitemporal vhr images based on
  deep kernel pca convolutional mapping network.
\newblock {\em IEEE Transactions on Cybernetics}, 52(11):12084--12098, 2022.

\bibitem{53}
Kunping Yang, Gui-Song Xia, Zicheng Liu, Bo~Du, Wen Yang, Marcello Pelillo, and
  Liangpei Zhang.
\newblock Asymmetric siamese networks for semantic change detection in aerial
  images.
\newblock {\em IEEE Transactions on Geoscience and Remote Sensing}, 60:1--18,
  2021.

\bibitem{90}
Meijuan Yang, Licheng Jiao, Fang Liu, Biao Hou, Shuyuan Yang, and Meng Jian.
\newblock Dpfl-nets: Deep pyramid feature learning networks for multiscale
  change detection.
\newblock {\em IEEE Transactions on Neural Networks and Learning Systems},
  33(11):6402--6416, 2021.

\bibitem{83}
Nabil Zerrouki, Fouzi Harrou, Ying Sun, and Lotfi Hocini.
\newblock A machine learning-based approach for land cover change detection
  using remote sensing and radiometric measurements.
\newblock {\em IEEE Sensors Journal}, 19(14):5843--5850, 2019.

\bibitem{76}
Chenxiao Zhang, Peng Yue, Deodato Tapete, Liangcun Jiang, Boyi Shangguan,
  Li~Huang, and Guangchao Liu.
\newblock A deeply supervised image fusion network for change detection in high
  resolution bi-temporal remote sensing images.
\newblock {\em ISPRS Journal of Photogrammetry and Remote Sensing},
  166:183--200, 2020.

\bibitem{85}
Han Zhang, Ian~J. Goodfellow, Dimitris~N. Metaxas, and Augustus Odena.
\newblock Self-attention generative adversarial networks.
\newblock In {\em Proceedings of the 36th International Conference on Machine
  Learning}, volume~97, pages 7354--7363, Long Beach, California, {USA}, 2019.
  {PMLR}.

\bibitem{68}
Hanyu Zhang, Qi~Wang, Ronghua Zhang, Xiuyan Li, Xiaojie Duan, Yukuan Sun,
  Jianming Wang, and Jiabin Jia.
\newblock Image reconstruction for electrical impedance tomography (eit) with
  improved wasserstein generative adversarial network (wgan).
\newblock {\em IEEE Sensors Journal}, 2022.

\bibitem{20}
Hao Zhang, Jiteng Yuan, Xin Tian, and Jiayi Ma.
\newblock Gan-fm: Infrared and visible image fusion using gan with full-scale
  skip connection and dual markovian discriminators.
\newblock {\em IEEE Transactions on Computational Imaging}, 7:1134--1147, 2021.

\bibitem{67}
Xiangmo Zhao, Pengpeng Sun, Zhigang Xu, Haigen Min, and Hongkai Yu.
\newblock Fusion of 3d lidar and camera data for object detection in autonomous
  vehicle applications.
\newblock {\em IEEE Sensors Journal}, 20(9):4901--4913, 2020.

\end{thebibliography}

\begin{IEEEbiography}[{\includegraphics[width=1in,height=1.25in,clip,keepaspectratio]{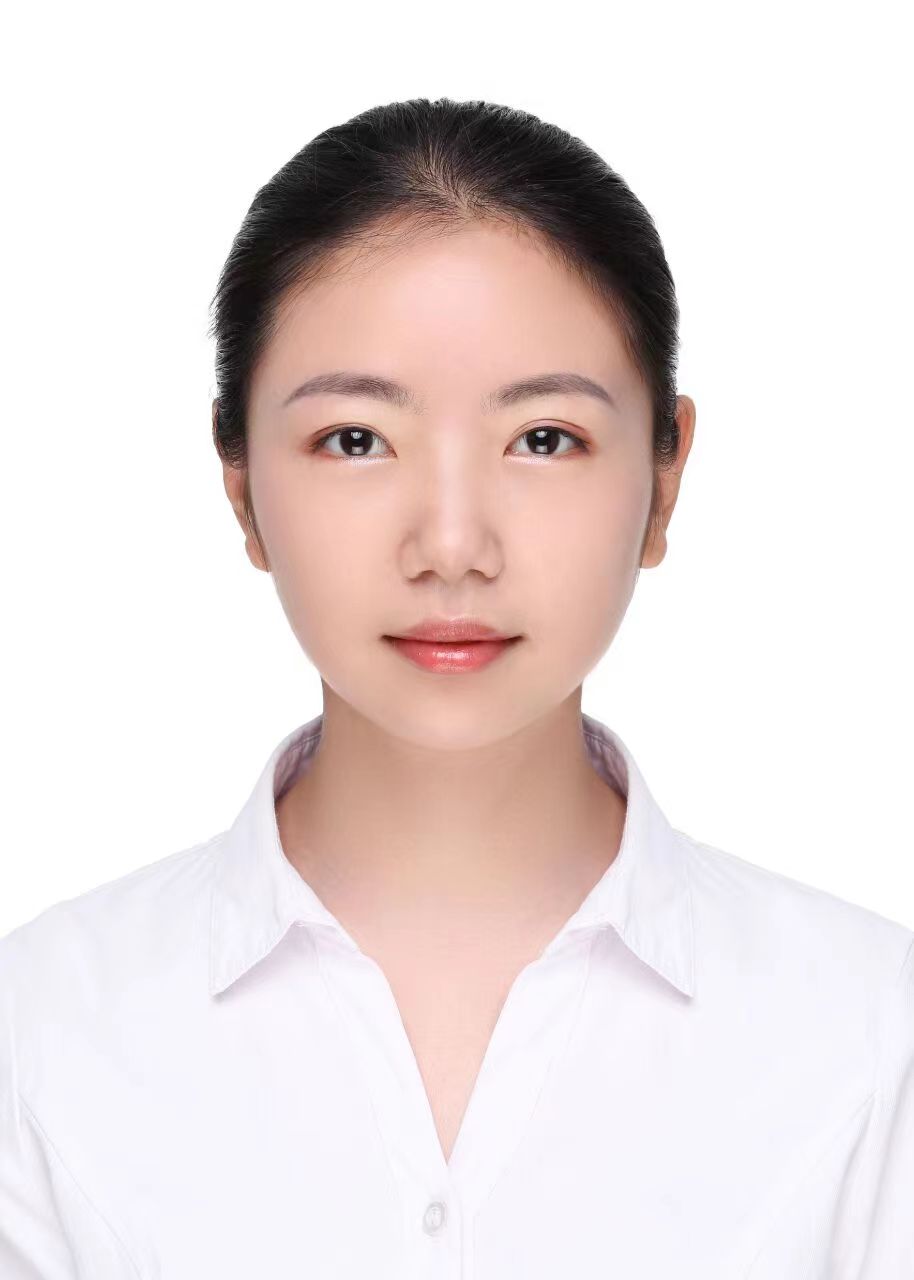}}]{LuYi Qiu} She received her Master's degree from the University of Electronic Science and Technology of China (UESTC) in 2023. She is a visiting scholar in the School of Electronic Information and Electrical Engineering at Shanghai Jiao Tong University. Her research interests include deep learning, data fusion analysis, and image processing.\end{IEEEbiography}

\begin{IEEEbiography}[{\includegraphics[width=1in,height=1.25in,clip,keepaspectratio]{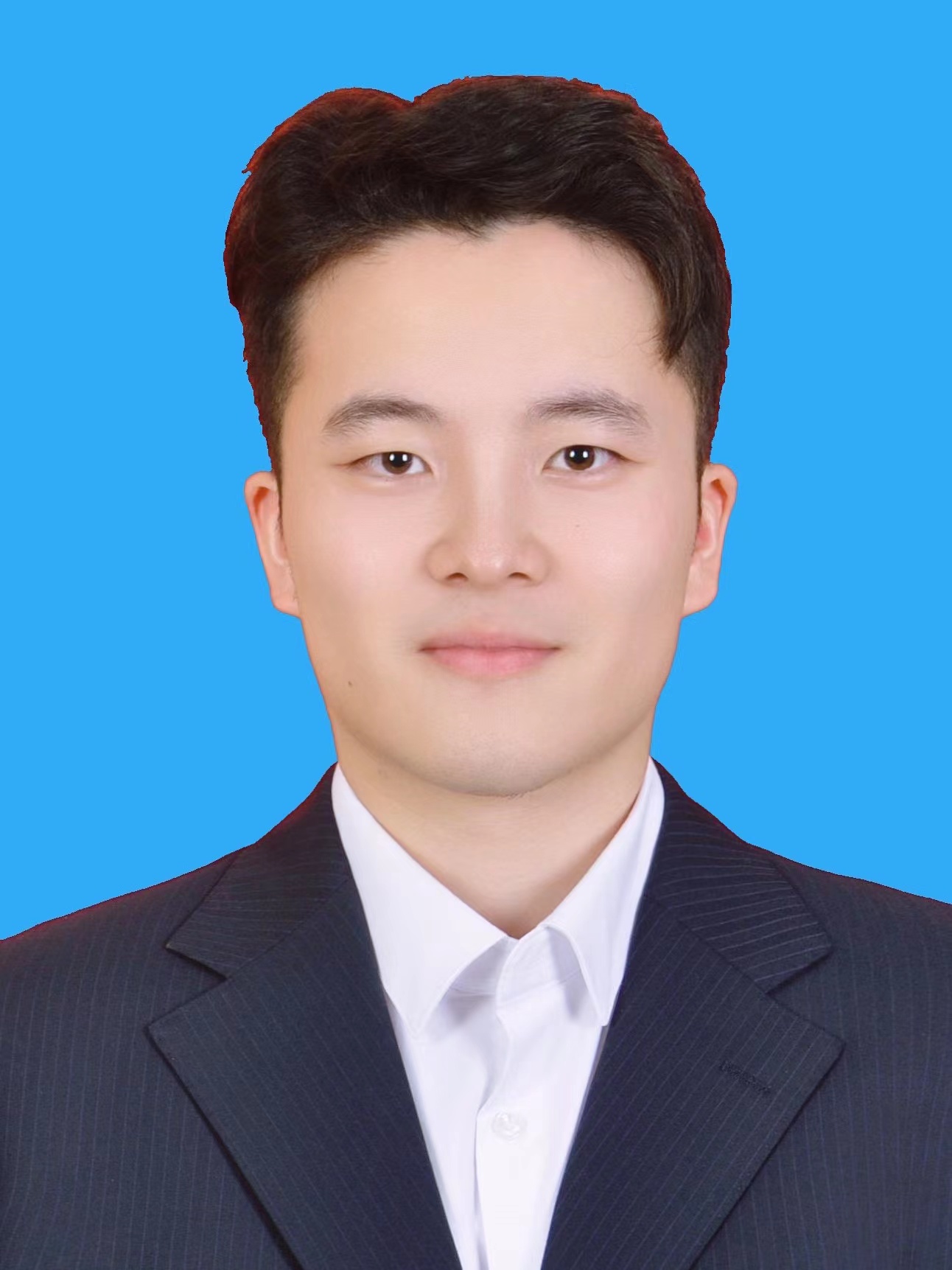}}]{XiaoFeng Zhang} (Student Member, IEEE) received his master's degree from Nanjing University of Posts and Telecommunications in 2021. He is now studying for his PhD at Shanghai Jiao Tong University, after researching deep learning, artificial intelligence, diffusion models, Segment anything model, Large-scale vision, and image restoration. \end{IEEEbiography}

\begin{IEEEbiography}[{\includegraphics[width=1in,height=1.25in,clip,keepaspectratio]{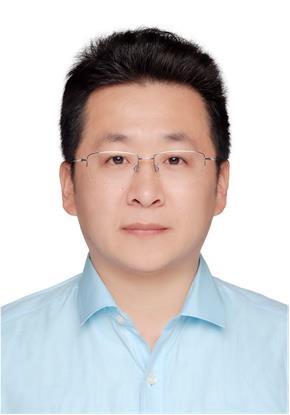}}]{ChaoChen Gu}(Member, IEEE) He graduated from the Department of Automation, Shanghai Jiao Tong University in 2013. He has long been engaged in the research of robot intelligent perception and autonomous control
In this field, he has completed a number of national, provincial and ministerial projects, and published more than 20 SCI/EI papers in this field
Tired of good research base and laboratory conditions. Program applicants have a solid professional background and interdisciplinary knowledge
Recognize the system, have a good research foundation for scene modeling and CAD technology, mechanism and robot control..\end{IEEEbiography}

\begin{IEEEbiography}[{\includegraphics[width=1in,height=1.25in,clip,keepaspectratio]{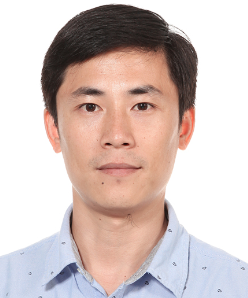}}]{ShanYing Zhu}(Member, IEEE)
He graduated from the Department of Automation, Shanghai Jiao Tong University in 2013. From 2013 to 2015, he conducted postdoctoral research at Nanyang Technological University in Singapore and Berkeley Education Consortium (BEARS). His research interests include Industrial Internet of Things, distributed estimation and optimization of network systems, multi-agent cooperative control, and energy management of micro-energy networks. He presided over 10 National Natural Science Foundation Outstanding Youth Fund, surface project, national key research and development plan sub-projects, and participated in 8 major/key projects of National Natural Science Foundation. He has published more than 80 papers and co-published an English monograph. TPC/IPC members who once served as multiple international conference, Invited Session/Local Arrangement/Publicity/Track Co - Chair, etc. He is currently a member of IEEE Industrial Informatics Technical Committee, a member of Youth Working Committee of Chinese Society of Automation, and a member of TCCT multi-agent Control Group.
 \end{IEEEbiography}

\newpage

\vfill

\end{document}